\documentclass[10pt,twocolumn,letterpaper]{article}

\usepackage{cvpr}      % To produce the REVIEW version
\usepackage{times}
\usepackage{epsfig}
\usepackage{graphicx}
\usepackage{amsmath}
\usepackage{amssymb}
\usepackage{color}
\usepackage{pifont}
\usepackage{booktabs}
\usepackage{multirow}
\newcommand{\tabincell}[2]{\begin{tabular}{@{}#1@{}}#2\end{tabular}}

\usepackage[hang,flushmargin]{footmisc}

\usepackage[pagebackref=true,breaklinks=true,letterpaper=true,colorlinks,bookmarks=false]{hyperref}

\def\cvprPaperID{92} % *** Enter the CVPR Paper ID here
\def\confName{CVPR}
\def\confYear{2024}

\begin{document}
	
	%%%%%%%%% TITLE
	\title{NTIRE 2024 Challenge on Stereo Image Super-Resolution: Methods and Results}
	\author{Longguang Wang, Yulan Guo$^\dag$, Juncheng Li, Hongda Liu, Yang Zhao, Yingqian Wang, Zhi Jin,\\ 
    Shuhang Gu, Radu Timofte, Davinci, Saining Zhang, Rongxin Liao, Ronghui Sheng, Feng Li,\\ 
    Huihui Bai, Wei Zhang, Runmin Cong, Yuqiang Yang, Zhiming Zhang, Jingjing Yang, Long Bao,\\ 
    Heng Sun, Kanglun Zhao, Enyuan Zhang, Huiyuan Fu, Huadong Ma, Yuanbo Zhou, Wei Deng,\\ 
    Xintao Qiu, Tao Wang, Qinquan Gao, Tong Tong,
    Yinghao Zhu, Yongpeng Li, Zhitao Chen,\\ 
    Xiujuan Lang, Kanghui Zhao, Bolin Zhu,  Wenbin Zou, Yunxiang Li, Qiaomu Wei, Tian Ye,\\ 
    Sixiang Chen,  Weijun Yuan, Zhan Li, Wenqin Kuang, Ruijin Guan,  Jian Wang, Yuqi Miao,\\ 
    Baiang Li, Kejie Zhao, Wenwu Luo, Jing Wu, Yunkai Zhang, Songyan Zhang, Jingyi Zhang,\\
    Junyao Gao, Xueqiang You, Yanhui Guo, Hao Xu, Sahaj K. Mistry, Aryan Shukla,\\ 
    Sourav Saini, Aashray Gupta, Vinit Jakhetiya, Sunil Jaiswal, Zhao Zhang, Bo Wang,\\ 
    Suiyi Zhao, Yan Luo, Yanyan Wei, Yihang Chen, Ruting Deng, Yifan Deng,\\Jingchao Wang, Zhijian Wu, Dingjiang Huang, Yun Ye}
	
	%\author{Longguang Wang$^{1}$, Yingqian Wang$^{1}$, Juncheng Li$^{2}$ Shuhang Gu$^{3}$, Radu Timofte$^{4}$, Yulan Guo$^{1*}$\\
	%	$^{1}$National University of Defense Technology~~~
	%	$^{2}$The Chinese University of Hong Kong\\
	%	$^{3}$The University of Sydney~~~~~~~~~~~~~~~~~~
	%	$^{4}$ETH Zürich\\
	%	{\tt\small \{wanglongguang15,yulan.guo\}@nudt.edu.cn}
	%}
	
	\maketitle

	%%%%%%%%% ABSTRACT
	\begin{abstract}
		This paper summarizes the 3rd NTIRE challenge on stereo image super-resolution (SR) with a focus on new solutions and results. The task of this challenge is to super-resolve a low-resolution stereo image pair to a high-resolution one with a magnification factor of $\times4$ under a limited computational budget. Compared with single image SR, the major challenge of this challenge lies in how to exploit additional information in another viewpoint and how to maintain stereo consistency in the results. This challenge has 2 tracks, including one track on bicubic degradation and one track on real degradations. In total, 108 and 70 participants were successfully registered for each track, respectively. In the test phase, 14 and 13 teams successfully submitted valid results with PSNR (RGB) scores better than the baseline. This challenge establishes a new benchmark for stereo image SR. 
	\end{abstract}
	
	%%%%%%%%% BODY TEXT
	\section{Introduction}
	\footnotetext{
		%\noindent *Longguang Wang, Yulan Guo, Juncheng Li, Hongda Liu, Yang Zhao, Yingqian Wang, Zhi Jin, Shuhang Gu, and Radu Timofte are the NTIRE 2024 challenge organizers, while the other authors participated in this challenge.	\\
		\noindent $^\dag$Corresponding author: Yulan Guo.	
		%L. Wang (wanglongguang15@nudt.edu.cn, National University of Defense Technology), Y. Guo (yulan.guo@nudt.edu.cn, National University of Defense Technology), Y. Wang, J. Li, S. Gu, and R. Timofte are the NTIRE 2022 organizers, while the other authors participated in the challenge.
		\\~~Section \ref{appendix} provides the authors and affiliations of each team.
		\\~~NTIRE 2024 webpage: \url{https://cvlai.net/ntire/2024/}
		\\~~Challenge webpage (Track 1): \url{https://codalab.lisn.upsaclay.fr/competitions/17245}
		\\~~Challenge webpage (Track 2): \url{https://codalab.lisn.upsaclay.fr/competitions/17246}
		\\~~Github: \url{https://github.com/The-Learning-And-Vision-Atelier-LAVA/Stereo-Image-SR/tree/NTIRE2024}
	}
    Recently, dual cameras have been increasingly popular in AR/VR, mobile phones, autonomous vehicles and robots to record and perceive the 3D environment. For higher perceptual quality and finer-grained parsing of the real world, increasing the resolution of stereo images is highly demanded. To this end, stereo image super-resolution (SR) has been introduced to reconstruct a high-resolution (HR) stereo image pair with finer details from a low-resolution (LR) one.
	
    Compared with a single image, stereo images can provide additional cues from a second viewpoint to better recover image details. However, since an object is projected onto different locations in left and right views, how to make full use of these cross-view information still remains challenging. On the one hand, stereo correspondence for objects at different depths can vary significantly. On the other hand, the occlusion between left and right views hinders correspondences to be incorporated.
	
    To develop and benchmark stereo SR methods, stereo image SR challenge was hosted in the NTIRE 2022 and NTIRE 2023 workshops \cite{Wang2022NTIRE,Wang2023NTIRE}. The objective of previous challenges is to minimize the distortion between super-resolved stereo images and the groundtruth under both bicubic and realistic degradations. However, the computational cost of the stereo SR methods is not fully considered, which hinders these methods to be deployed on resource-limited devices.
    %This challenge, employed the Flick1024 dataset \cite{Wang2019Flickr1024} and widely-applied bicubic degradation to synthesize LR stereo images. Besides, However, degradations in real-world scenarios are more complicated than the bicubic one. In addition, the perceptual quality and the stereo consistency are also critical to the visual effects of stereo images.     
	
    Succeeding the previous years, NTIRE 2024 Stereo Image SR Challenge presents two competition tracks. These two tracks are inherited from the NTIRE 2023 challenge with an additional constraint of computational complexity. Specifically, both the memory and computational cost are  considered for real-world applications.
    %Track 1 is inherited from the NTIRE 2023 challenge, focusing on bicubic degradation and restoration accuracy in terms of PSNR. Track 2 also adopts bicubic degradation to synthesize LR images but focus on restoration accuracy in terms of LPIPS. In track 3, complicated degradations including blur, noise, downsampling, and compression are used for LR image synthesis, with PSNR being employed to measure restoration accuracy. 
	
 This challenge is one of the NTIRE 2024 Workshop associated challenges on: dense and non-homogeneous dehazing~\cite{ntire2024dehazing}, night photography rendering~\cite{ntire2024night}, blind compressed image enhancement~\cite{ntire2024compressed}, shadow removal~\cite{ntire2024shadow}, efficient super resolution~\cite{ntire2024efficientsr}, image super resolution ($\times$4)~\cite{ntire2024srx4}, light field image super-resolution~\cite{ntire2024lightfield}, stereo image super-resolution~\cite{ntire2024stereosr}, HR depth from images of specular and transparent surfaces~\cite{ntire2024depth}, bracketing image restoration and enhancement~\cite{ntire2024bracketing}, portrait quality assessment~\cite{ntire2024QA_portrait}, quality assessment for AI-generated content~\cite{ntire2024QA_AI}, restore any image model (RAIM) in the wild~\cite{ntire2024raim}, RAW image super-resolution~\cite{ntire2024rawsr}, short-form UGC video quality assessment~\cite{ntire2024QA_UGC}, low light enhancement~\cite{ntire2024lowlight}, and RAW burst alignment and ISP challenge.

    \section{{Related Work}}\label{SecRelatedWork}
    In this section, we briefly review recent advances in single image SR and stereo image SR.
	
    \subsection{Single Image SR}
    In the last decade, learning-based approaches have dominated the area of single image SR \cite{Deng2024Efficient,Zhou2023Memory,Chen2023Dual,Wang2021Unsupervised,Li2024Asystematic}. Dong \textit{et al.} \cite{Dong2014Learning} proposed the first CNN-based SR model (\textit{i.e.}, SRCNN) to learn an LR-to-HR mapping. Following SRCNN, early methods focus on developing larger and more effective network architectures to achieve higher SR performance \cite{Kim2016Accurate,Lim2017Enhanced,Zhang2018Image}. Specifically, Zhang \textit{et al.} \cite{Zhang2018Residual} proposed a residual dense network (\textit{i.e.}, RDN) to fully use hierarchical features by combining residual connection \cite{He2016Deep} with dense connection \cite{Huang2017Densely}. Subsequently, Li \textit{et al.} \cite{Li2018Multi} suggested to use image features at different scales for single image SR, and proposed a multi-scale residual network (\textit{i.e.}, MSRN). Dai \textit{et al.} \cite{Dai2019Second} proposed a second-order attention network (\textit{i.e.}, SAN) for more powerful feature correlation learning, which achieves superior performance. Recently, the efficiency of SR models has drawn increasing interests, with numerous lightweight network architectures being developed \cite{Song2021Addersr,Wang2021Learning,Wang2021Exploring}. For example, distillation blocks were proposed for feature learning in IDN \cite{Hui2018Fast}. Then, a cascading mechanism was introduced to encourage efficient feature reuse in CARN \cite{Ahn2018Fast}. Different from these manually designed networks, Chu \emph{et al.} \cite{Chu2020Fast} developed a compact architecture using neural architecture search (NAS). 
 
    Inspired by the great success of Transformer in computer vision, Transformer has been widely studied to promote single image SR. Liang \textit{et al.} \cite{Liang2021SwinIR} designed a SwinIR model for image restoration by applying Swin Transformer \cite{Liu2021Swin}. Lu \textit{et al.} \cite{Lu2022Transformer} proposed an effective super-resolution Transformer (\textit{i.e.}, ESRT) for single image SR, which introduced a lightweight Transformer and feature separation strategy to reduce GPU memory consumption. Zamir \textit{et al.} \cite{Zamir2022Restormer} proposed an encoder-decoder Transformer (\textit{i.e.}, Restormer) for image restoration with multi-scale local-global representation learning. %These models not only produce promising results on single image SR, but also perform well on other image restoration tasks. More detailed and advanced methods can refer to recent surveys \cite{Yang2019Deep,Wang2019Deep,Li2021beginner}.
	
	%The aforementioned methods mainly focus on reconstructing HR images with high fidelity (evaluated by PSNR and SSIM metrics) under the standard bicubic degradation. Apart from these methods, there are many methods that studied various challenging issues in single image SR such as SR under multiple degradation \cite{Zhang2017Learninga,Gu2019Blind,Wang2021Learning,liang2021mutual}, generating visually pleasant (may not faithful) details \cite{Wang2018ESRGAN}, SR with non-integral factors \cite{Hu2019Meta,chen2021learning,Wang2021Learning}, and efficient inference \cite{Kong2021ClassSR, Wang2021Exploring} etc. Readers can refer to recent surveys \cite{Wang2020Deep,li2021beginner,liu2021blind} to learn more details about single image SR.
	
    \subsection{Stereo Image SR}
    Jeon \textit{et al.} \cite{Jeon2018Enhancing} developed the first learning-based stereo image SR method termed StereoSR. This method incorporates cross-view information by concatenating the left image and a stack of right images with different pre-defined shifts. Later, Wang \textit{et al.} \cite{Wang2019Learning,Wang2020Parallax} developed PASSRnet by introducing a parallax attention module (PAM) to capture stereo correspondence along the epipolar line. Inspired by PASSRnet, Song \textit{et al.} \cite{Song2020Stereoscopic} further combined self-attention with parallax attention to better model global correspondence. Wang \textit{et al.} \cite{Wang2021Symmetric} introduced a Siamese network with a bi-directional parallax attention module to simultaneously super-resolve left and right images in a symmetric manner. Guo \textit{et al.}~\cite{Guo2023Pft} proposed a new Transformer-based parallax fusion model called Parallax Fusion Transformer.

    Instead of using parallax-attention mechanism, several efforts have also been made to employ stereo matching approach to capture stereo correspondence. Yan \textit{et al.} \cite{Yan2020Disparity} proposed a domain adaptive stereo SR network (DASSR) to incorporate cross-view information through explicit disparity estimation using a pre-trained stereo matching network. Dai \textit{et al.} \cite{Dai2021Feedback} proposed a feedback network to alternately solve disparity estimation and stereo image SR in a recurrent manner. Wan \textit{et al.} \cite{Wan2023Multi} proposed a multi-stage network to progressively obtain cross-view features and an edge-guided supplementary branch to refine the cross-view features.

    In the NTIRE 2022 Stereo Image SR Challenge, the champion team developed NAFSSR network \cite{Chu2022NAFSSR} by using nonlinear activation-free network (NAFNet) for feature extraction and PAM for cross-view information interaction. In the NTIRE 2023 Stereo Image SR Challenge, the champion team of track 1 developed a Hybrid Transformer and CNN Attention Network (HTCAN), which employs a Transformer-based network for single image enhancement and a CNN-based network for stereo information aggregation. The champion team of track 2 proposed a SwinFIRSSR by using Swin Transformer~\cite{Liu2021Swin} and fast Fourier convolution~\cite{Chi2020Fast}. The champion team of track 3 combined NAFSSR \cite{chen2022simple} with LTE \cite{Chen2021Learning} and proposed LTFSSR.
    
    %The runner-up team proposed a SwiniPASSR network by combining the Swin Transformer with PAM. The second runner-up team also adopted a Transformer-based network termed SSRFormer with a Siamese structure.
	
    \section{NTIRE 2024 Challenge}
    The objectives of the NTIRE 2024 challenge on example-based stereo image SR are: (i) to gauge and push the state-of-the-art in SR under given computational constraints; and (ii) to compare different solutions.
	
    \subsection{Dataset}
    \noindent \textbf{Training Set.}
    The training set of the Flickr1024 dataset \cite{Wang2019Flickr1024} (with 800 images) is used as the training set of this challenge. Both original HR images and their LR versions will be released. The participants can use these HR images as ground-truth to train their models. 
	
	\noindent \textbf{Validation Set.} 
	The validation set of the Flickr1024 dataset (with 112 images) is used as the validation set of this challenge.
	Similar to the training set, both HR and LR images in the validation set are provided. The participants can download the validation set to evaluate the performance of their developed models by comparing their super-resolved images with the HR ground-truth images. Note that the validation set should be used for validation purposes only but cannot be used as additional training data. %The participants are encouraged to write papers to describe their methods and use the released validation set for performance evaluation.
	
	\noindent \textbf{Test Set.}
	To rank the submitted models, a test set consisting of 100 stereo images is provided. Unlike the training and validation sets, only LR images will be released for the test set. The participants must apply their models to the released LR stereo images and submit their super-resolved images to the server. 
	It should be noted that the images in the test set (even the LR versions) cannot be used for training. 
	
	\subsection{Tracks}
	\noindent{\textbf{$\bullet$ Track 1: Constrained SR \& Bicubic Degradation}}
	\vspace{0.1cm}
	
    In this track, bicubic degradation (Matlab $imresize$ function in bicubic mode) is used to generate LR images:
    \begin{equation}
    	I^{LR}=I^{HR}\downarrow_s,
    \end{equation}
    where $I^{LR}$ and $I^{HR}$ are LR and HR images, $\downarrow_s$ represents bicubic downsampling with scale factor $s$.

	\vspace{0.1cm}
	\noindent{\textbf{$\bullet$ Track 2: Constrained SR \& Realistic Degradation}}
	\vspace{0.1cm}

    In this track, a realistic degradation model consisting of blur, downsampling, noise, and compression is adopted to synthesize LR images:
    \begin{equation}
    I^{LR}=\mathcal{C}\left(\left(I^{HR}\otimes{k}\right)\downarrow_s+n\right),
    \end{equation}
    where $k$ is the blur kernel, $n$ is additive Gaussian noise, and $\mathcal{C}$ represents JPEG compression. 
 
    In these two tracks, the model size (\textit{i.e.}, number of parameters) is restricted to 1 MB, and the computational complexity (\emph{i.e.}, number of MACs) is restricted to 400 G (a stereo image pair of size $320\times180$).  Peak signal-to-noise ratio (PSNR) and structural similarity (SSIM) are used as metrics for performance evaluation. The average results of left and right views over all of the test scenes are reported. Note that only PSNR (RGB) is used for ranking.

    \begin{table*}[t]
		\caption{NTIRE 2024 Stereo Image SR Challenge (Track 1) results, rankings, and details from the fact sheets. Note that, PSNR (RGB) is used for the ranking. ``Transf'' denotes Transformer and ``PAM'' denotes parallax attention mechanism.}
		\label{tab1}
		\centering
		\scriptsize
		\setlength{\tabcolsep}{1mm}{
			\begin{tabular}{cllcccccccc}
				\hline
				Rank &Team
				& Authors
                & PSNR (RGB)
				& Architecture
				& Disparity
				& Ensemble
				\tabularnewline
				\hline
                1 & \tabincell{l}{Davinci} & \tabincell{l}{Davinci and S. Zhang} & 23.6503 & CNN+Transf & PAM & Data+Feature
			\tabularnewline
                2 & \tabincell{l}{HiSSR} & \tabincell{l}{R. Liao, R. Sheng, F. Li, H. Bai, R. Cong, and W. Zhang} & 23.6105 & CNN & PAM & N.A.
                \tabularnewline
                3 & \tabincell{l}{MiVideoSR} & \tabincell{l}{Y. Yang, Z. Zhang, J. Yang, L. Bao, and H. Sun} & 23.6070 & CNN+Transf & PAM & Data
			\tabularnewline
                4 & \tabincell{l}{webbzhou} & \tabincell{l}{Y. Zhou, W. Deng, X. Qiu, T. Wang, Q. Gao, and T. Tong} & 23.5941 & CNN & PAM & N.A.
			\tabularnewline
                5 & \tabincell{l}{Qi5} & \tabincell{l}{Y. Zhu and Y. Li} & 23.5896 & CNN+Transf & PAM & N.A.
			\tabularnewline
                6 & \tabincell{l}{WITAILab} & \tabincell{l}{Z. Chen, X. Lang, K. Zhao, and B. Zhu} & 23.5725 & CNN+Transf & PAM & N.A.
			\tabularnewline
                7 & \tabincell{l}{Giantpandacv} & \tabincell{l}{W. Zou, Y. Li, Q. Wei, T. Ye, and S. Chen} & 23.5271 & CNN & PAM & N.A.
			\tabularnewline
                8 & \tabincell{l}{JNU\_620} & \tabincell{l}{W. Yuan, Z. Li, W. Kuang, and R. Guan} & 23.4851 & CNN & PAM & N.A.
			\tabularnewline
                9 & \tabincell{l}{GoodGame} & \tabincell{l}{J. Wang, Y. Miao, B. Li, and K. Zhao} & 23.4598 & CNN & PAM & N.A.
			\tabularnewline
                10 & \tabincell{l}{Fly\_Flag} & \tabincell{l}{W. Luo, and J. Wu} & 23.4510 & CNN+Transf & PAM & Data
			\tabularnewline
                11 & \tabincell{l}{Mishka} & \tabincell{l}{Y. Zhang, B. Li, S. Zhang, J. Zhang, J. Gao, and X. You} & 23.4270 & CNN+Mamba & PAM & N.A.
			\tabularnewline
                12 & \tabincell{l}{LightSSR} & \tabincell{l}{Y. Guo and H. Xu} & 23.3888 & CNN & N.A. & N.A.
			\tabularnewline
                13 & \tabincell{l}{DVision} & \tabincell{l}{S. Mistry, A. Shukla, S. Saini, A. Gupta, V. Jakhetiya, and  S. Jaiswal} & 23.1895 & CNN & PAM & N.A.
			\tabularnewline
                14 & \tabincell{l}{LVGroup\_HFUT} & \tabincell{l}{Z. Zhang, B. Wang, S. Zhao, Y. Luo, and Y. Wei} & 23.0977 & CNN & N.A. & Data
			\tabularnewline

				\hline
		\end{tabular}}
    \end{table*}

    \begin{table*}[t]
		\caption{NTIRE 2024 Stereo Image SR Challenge (Track 2) results, rankings, and details from the fact sheets. Note that, PSNR (RGB) is used for the ranking. ``Transf'' denotes Transformer and ``PAM'' denotes parallax attention mechanism.}
		\label{tab2}
		\centering
		\scriptsize
		\setlength{\tabcolsep}{1.5mm}{
			\begin{tabular}{cllcccccccc}
				\hline
				Rank &Team
				& Authors
				& PSNR (RGB)
				& Architecture
				& Disparity
				& Ensemble
				\tabularnewline
				\hline
                1 & \tabincell{l}{Davinci} & \tabincell{l}{Davinci and S. Zhang} & 21.8724 & CNN+Transf. & PAM & Data+Feature
			\tabularnewline
                2 & \tabincell{l}{MiVideoSR} & \tabincell{l}{Y. Yang, Z. Zhang, J. Yang, L. Bao, and H. Sun} & 21.6983 & CNN+Transf & PAM & Data
                \tabularnewline
                3 & \tabincell{l}{BUPTMM} & \tabincell{l}{K. Zhao, E. Zhang, H. Fu, and H. Ma} & 21.6702 & CNN & PAM & N.A.
			\tabularnewline
                4 & \tabincell{l}{webbzhou} & \tabincell{l}{Y. Zhou, W. Deng, X. Qiu, T. Wang, Q. Gao, and T. Tong} & 21.6691 & CNN & PAM & N.A.
			\tabularnewline
                5 & \tabincell{l}{JNU\_620} & \tabincell{l}{W. Yuan, Z. Li, W. Kuang, and R. Guan} & 21.5935 & CNN & PAM & N.A.
			\tabularnewline
                6 & \tabincell{l}{Liz620} & \tabincell{l}{Y. Chen, R. Deng, and Y. Deng} & 21.5655 & CNN & PAM & N.A.
			\tabularnewline
                7 & \tabincell{l}{Mishka} & \tabincell{l}{Y. Zhang, B. Li, S. Zhang, J. Zhang, J. Gao, and X. You} & 21.5313 & CNN+Mamba & PAM & N.A.
			\tabularnewline
                8 & \tabincell{l}{ECNU-IDEALab} & \tabincell{l}{J. Wang, Z. Wu, and D. Huang} & 21.5238 & CNN+Transf & PAM & N.A.
			\tabularnewline
                9 & \tabincell{l}{Giantpandacv} & \tabincell{l}{W. Zou, Y. Li, Q. Wei, T. Ye, and S. Chen} & 21.4970 & CNN & PAM & N.A.
			\tabularnewline
                10 & \tabincell{l}{HiYun} & \tabincell{l}{Y. Ye} & 21.1994 & CNN & PAM & Data
			\tabularnewline
                11 & \tabincell{l}{GoodGame} & \tabincell{l}{J. Wang, Y. Miao, B. Li, and K. Zhao} & 20.7642 & CNN & PAM & N.A.
			\tabularnewline
                12 & \tabincell{l}{Fly\_Flag} & \tabincell{l}{W. Luo, and J. Wu} & 20.7518 & CNN+Transf & PAM & Data
			\tabularnewline
                13 & \tabincell{l}{LVGroup\_HFUT} & \tabincell{l}{Z. Zhang, B. Wang, S. Zhao, Y. Luo, and Y. Wei} & 20.6167 & CNN & N.A. & Data
			\tabularnewline
   
				\hline
		\end{tabular}}
    \end{table*}

	\subsection{Challenge Phases}
	\noindent{\textbf{Development Phase.}} 
	The participants were provided with pairs of LR and HR training images and LR validation images of the Flickr1024 dataset. The participants had the opportunity to test their solutions on the LR validation images and to receive immediate feedback by uploading their results to the server. A validation leaderboard is available online.
	
    \noindent{\textbf{Testing Phase.}} 
    The participants were provided with the LR test images and were asked to submit their super-resolved images, codes, and a fact sheet for their methods before the challenge deadline. After the end of the challenge, the final results were released to the participants.
	
    \section{{Challenge Results}}
    
    \subsection{Track 1: Fidelity \& Bicubic Degradation}

    {Among the 108 registered participants, 14 teams successfully participated the final phase and submitted their results, codes, and fact sheets. Table \ref{tab1} reports the final test results, rankings of the challenge, and major details from the fact sheets of 14 teams. These methods are briefly described in Section \ref{sec4} and the team members are listed in Section \ref{appendix}.}

    %It can be observed that the top 5 teams successfully outperform the winner method in NTIRE 2022 (\emph{i.e.}, NAFSSR \cite{Chu2022NAFSSR} 23.7873), further boosting the performance of stereo image SR. Moreover, the accuracy of the top 2 methods are very close with a minor PSNR difference of 0.005. In addition, although the SRC-B team produces inferior PSNR results than the top 2 teams, it achieves the highest SSIM score of 0.7400.

    \subsection{Track 2: Fidelity \& Realistic Degradation}

    {Among the 70 registered participants, 13 teams successfully participated the final phase and submitted their results, codes, and fact sheets. Table \ref{tab2} reports the final test results, rankings of the challenge, and major details from the fact sheets of 13 teams. These methods are briefly described in Section \ref{sec4} and the team members are listed in Section \ref{appendix}.}

    %As we can see, all methods suffer a notable performance drop on realistic degradations as compared to the standard bicubic one. In addition, although the Team OV team produces inferior PSNR results than the IPIU team, it achieves the highest SSIM score of 0.6030.

    \subsection{Summary}
 
    \textbf{Architectures and main ideas.}
    All the proposed methods are based on deep learning techniques. Transformers and the winner method in the NTIRE 2022 challenge (\emph{i.e.}, NAFSSR) are widely used as the basic architecture.  To exploit cross-view information, the idea of parallax-attention mechanism (PAM) are widely adopted in most solutions to capture stereo correspondence.
	
    \textbf{Data Augmentation.}
    Widely applied data augmentation approaches such as random flipping and RGB channel shuffling are used for most solutions. In addition, random horizontal shifting, Mixup, CutMix, and CutMixup are also used in several solutions and help to achieve superior performance. 
	
    \textbf{Ensembles and Fusion.}
    Due to the constraints in computational complexity, ensemble strategy (data ensemble and model ensemble) is only adopted in a few solutions. Several solutions employ a limited number of transformed inputs for enhanced prediction. In addition, the champion solution in track 1 employs model exponential moving average for model ensemble without additional overhead during the inference phase.
	
	%\noindent \textbf{Exploitation of stereo correspondence.}
	%disparity estimation or PAM or other techniques to make use of disparity information
	
	\textbf{Conclusions.}
	By analyzing the settings, the proposed methods and their results, it can be concluded that: 
	1) The proposed methods strike better balance between accuracy and efficiency.
	2) With recent renaissance of CNNs (\emph{e.g.}, NAFNet), Transformers and CNNs are comparably popular in this challenge and produce competitive performance.
	3) Cross-view information lying at varying disparities is critical to the stereo image SR task and helps to achieve higher performance.
	4) To meet the efficiency requirements of the challenge, techniques like depth-wise convolution are widely applied and produce promising results.
        5) One recent remarkable technique (\emph{i.e.}, Mamba) has been introduced to achieve efficient image SR and produces promising results.

    \section{Challenge Methods and Teams}
    \label{sec4}

    \subsection{Davinci - Track 1\textcolor[RGB]{255,215,0}{$^\bigstar$}, Track 2\textcolor[RGB]{255,215,0}{$^\bigstar$}}

\begin{figure*}[t]
		\centering
		\includegraphics[width=1.0\linewidth]{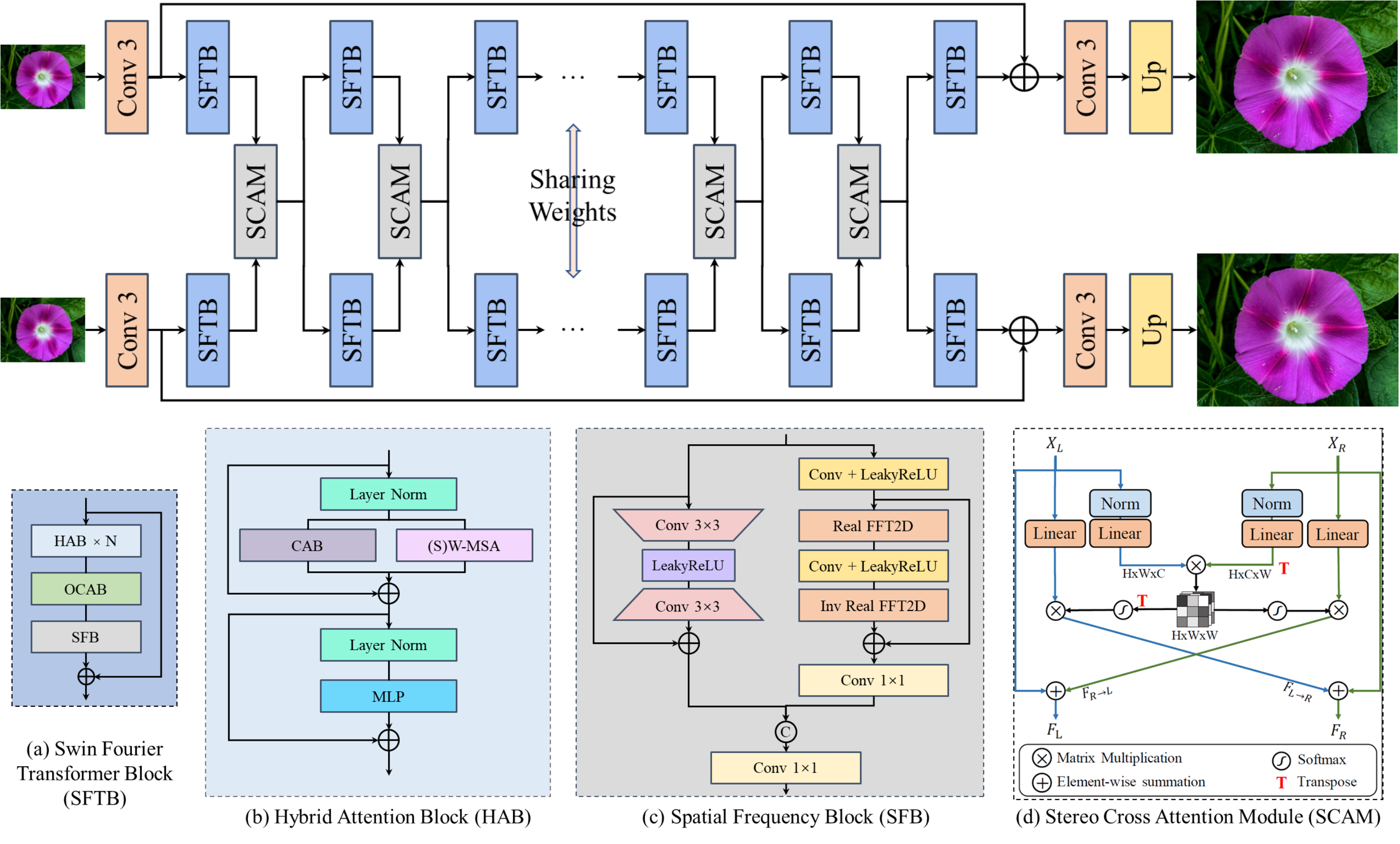}
		\caption{Davinci: The structure of the proposed SwinFIRSSR.}
		\label{fig:Davinci}
    \end{figure*}

    Inspired by SwinFIR~\cite{Zhang2022Swinfir}, HAT~\cite{Chen2023Activating}, and NAFSSR~\cite{Chu2022NAFSSR}, they proposed SwinFIRSSR using Swin transformer~\cite{Liu2021Swin} and fast fourier convolution~\cite{Chi2020Fast}, as shown in Fig.~\ref{fig:Davinci}. HAT employs Residual Hybrid Attention Group (RHAG) to activate more pixel in image SR transformer to improve the performance. Each RHAG contains $N$ hybrid attention blocks (HAB), an overlapping cross-attention block (OCAB) and a 3$\times$3 convolutional layer. They replaced the convolution (3$\times$3) with fast fourier convolution and a residual module to fuse global and local features, namely Spatial-frequency Block (SFB), to improve the representation ability of the model. They also followed NAFSSR to fuse left/right features using stereo cross-attention module (SCAM). 

    During the training phase, HR images were cropped to 128$\times$384 sub-images. The Adam~\cite{Kingma2015Adam} optimizer with default parameters and the Charbonnier L1 loss~\cite{Lai2017Deep} were employed for training. The initial learning rate was set to $2\times10^{-4}$ and decayed at 600,000, 650,000, 700,000, 750,000 iterations. The batch size was 4 and patch size was 32$\times$96. The models were implemented usingPyTorch 1.8.1, NVIDIA A6000 GPU with CUDA11.1. Random horizontal flip, vertical flip, rotation, RGB perm and mixup~\cite{yoo2020rethinking} were adopted for data augmentation.
    
\begin{figure}[t]
        \centering
        \includegraphics[width=1.0\linewidth]{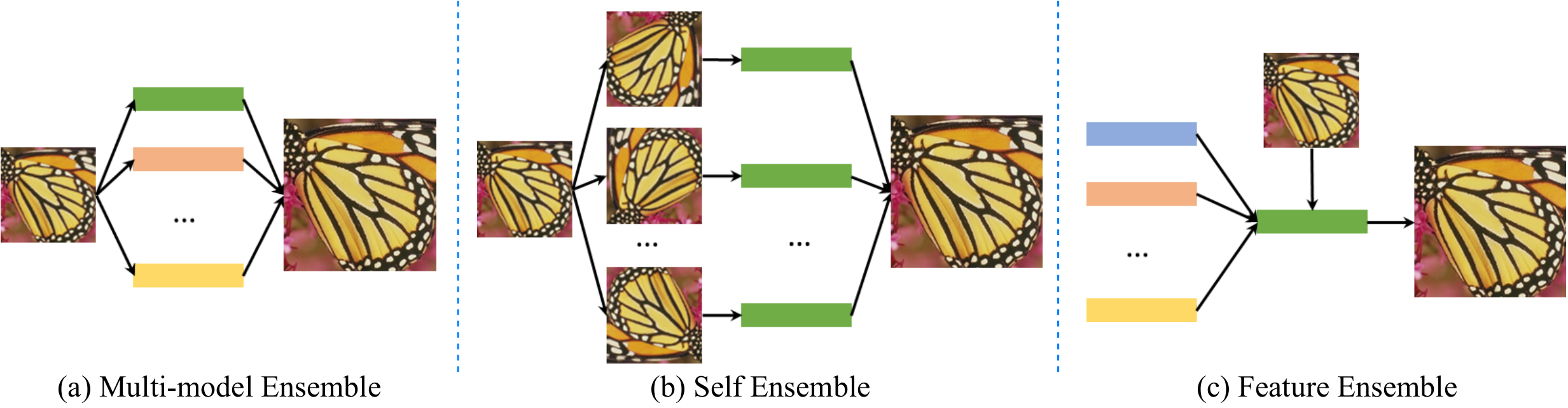}
        \caption{Comparison between different ensemble strategies. (a) Multi-model ensemble. (b) Data ensemble. (c) The proposed model ensemble. Rectangles with different colors represent different model parameters.} 
        \label{fig:FeatureEnsemble}
    \end{figure}

    Inspired by~\cite{Timofte2016Seven}, model ensemble was employed to improve the performance. Widely-applied multi-model and data ensemble strategy inevitably introduces additional computational overhead. To remedy this, they proposed to conduct model ensemble from the perspective of network parameters, as illustrated in Fig.~\ref{fig:FeatureEnsemble}. Specifically, the parameters of several models are aggregated as:
    \begin{align}
    	SwinFIRSSR(\theta) = \sum_{i=1}^n {SwinFIRSSR(\theta)^i * \alpha^i},
    \end{align}
    where $\theta$ denotes the parameter sets of SwinFIRSSR, $n$ is the numbers of models. $\alpha$ is the weight of each model and the $\alpha = \frac{1}{n}$ in our solution. 

    \subsection{HiSSR - Track 1\textcolor[RGB]{192,192,192}{$^\bigstar$}}

\begin{figure*}[t]
        \centering
        \includegraphics[width=1\textwidth]{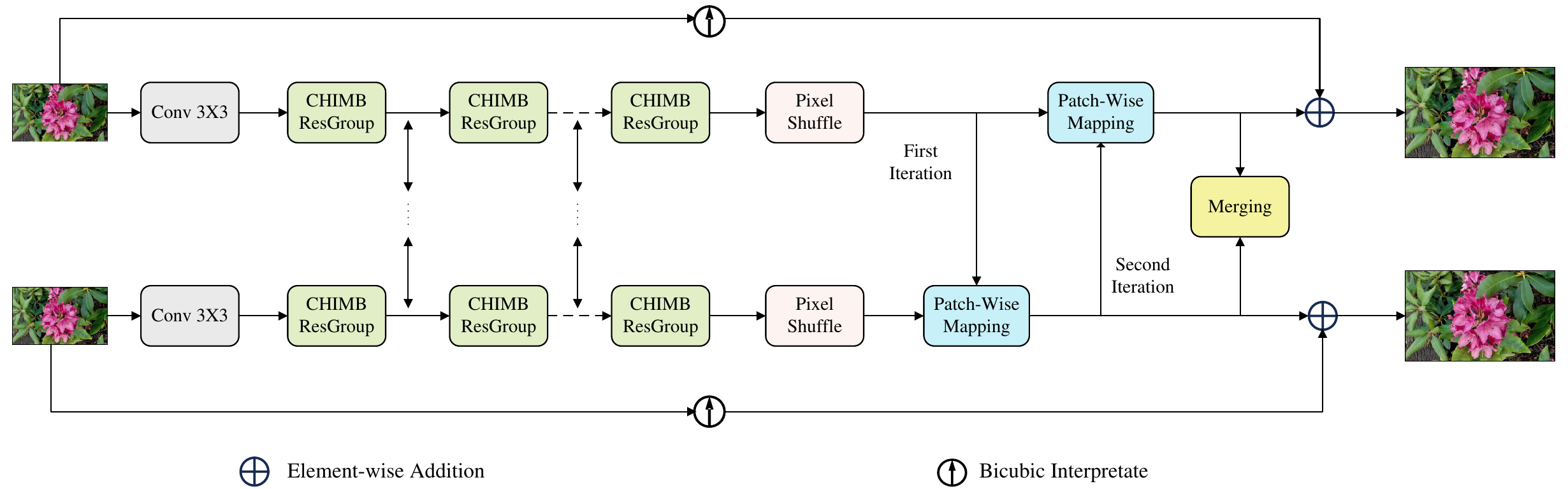}
        \caption{HiSSR: The structure of the proposed RISSR.}
        \label{fig:HiSSR}
    \end{figure*}

    Figure \ref{fig:HiSSR} depicts the network structure of the proposed method. Specifically, CVHSSR\cite{Zou2023Cross} is used as the backbone, which consists of a cross-hierarchy information mining block (CVIM) and a cross-hierarchy information mining block (CHIMB). CHIMB leverages both spatial and channel-wise attention mechanisms to model information at different levels within a single view. Meanwhile, CVIM integrates similar information across different views through cross-view attention. They employed CHVIMB to extract both global and local information from a single view, while CVIM is utilized to integrate analogous information across each views. Following CFSR\cite{Wu2024Transforming}, they further reparameterized the depth-wise convolution in the Information Refinement Feedforward Block in CHIMB. To further enhance the model's performance, they introduced an iterative interaction mechanism, leveraging the restoration results as reference images to enhance the performance.
        
    \textbf{Training Settings.}
    The number of CHIMB in RIISSR is set to 20, while the channels of all the convolutional layers are set to 48. Residual connections are inserted between every four blocks. The number of parameters in the model is 0.918M, and the MACs are 235.28G for a \(320 \times 180\) stereo image pair. 
    During training, HR images were cropped into (30 $\times$90) patches with a stride of 10. Random horizontal, flips, rotations, mixup and RGB channel shuffle were adopted for data augmentation. The AdamW with \(\beta_1\) = 0.9 and \(\beta_2\) = 0.99 was used for optimization. The learning rate was initialized to ($1 \times 10^{-3}$) and decreased using the multi-step strategy with \(\gamma\) set to 0.5. The proposed model was trained for ($2.4 \times 10^5$) iterations with a batch size of 24. A $2\times$ SR model was first trained from scratch, which was then used to initialize the $4\times$ SR model. 

   \subsection{MiVideoSR - Track 1\textcolor[RGB]{184,115,51}{$^\bigstar$}, Track 2\textcolor[RGB]{192,192,192}{$^\bigstar$}}

\begin{figure*}[t]
        \centering
        \includegraphics[width=1\textwidth]{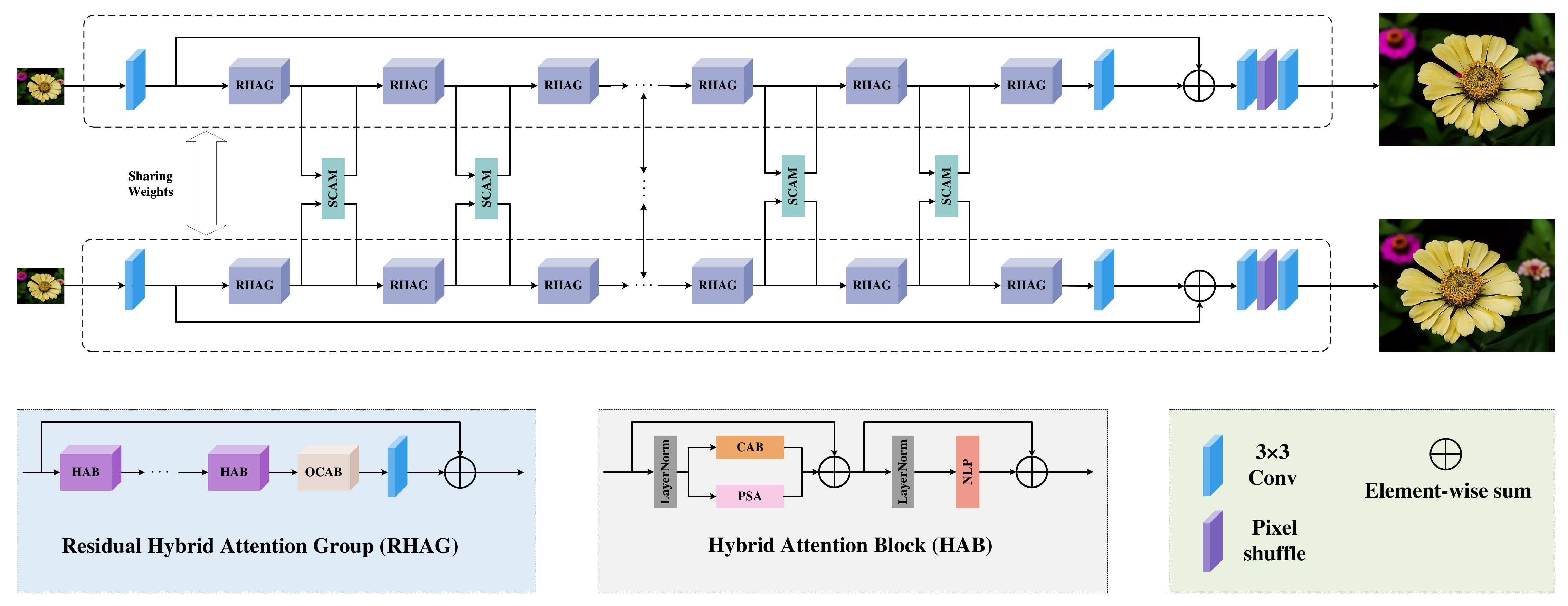} 
        \caption{MiVideoSR: The structure of the proposed HCASSR.}
        \label{fig:MiVideoSR}
    \end{figure*}

    Transformer-like methods have achieved advanced performance on low-level tasks. Motivated by NAFSSR\cite{Chu2022NAFSSR}, HAT\cite{Chen2023Activating} and SRFormer\cite{Zhou2023Srformer}, this team proposed a HCASSR by plugging the SCAM of NAFSSR into the HAT to aggregate features from two views, as illustrated in Fig.~\ref{fig:MiVideoSR}. In addition, to improve the performance and efficiency of the model, they replaced the self-attention module with permuted self-attention (PSA)\cite{Zhou2023Srformer} to transfer the spatial information to the channel dimension.

    \textbf{Training Settings.}
    The proposed model was first trained with a Charbonnier loss using an Adam optimizer and stopped after 400k iterations. Then, the resultant model was fine-tuned with the MSE loss. The batch size was set to $16$ and the patch size was first set to $96 \times 96$ and then enlarged to $192 \times 192$ for fine-tuning. The learning rate was initialized as $5 \times {10}^{-4}$ and updated using a cosine annealing strategy. Data augmentation was performed through horizontal/vertical flipping, RGB channel random shuffing, and Mixup operations.
    
    \textbf{Data Ensemble.}
    Due to the limitation of the number of parameters, model ensemble strategy was abandoned. Besides, due to the limitation of the computational complexity, only horizontal flipping and vertical flipping were used to generate three images for data ensemble.
%    three methods including the original image (the other two are horizontal flipping and vertical flipping).

   \subsection{BUPTMM - Track 2\textcolor[RGB]{184,115,51}{$^\bigstar$}}

\begin{figure*}[t]
        \centering
        \includegraphics[width=0.9\textwidth]{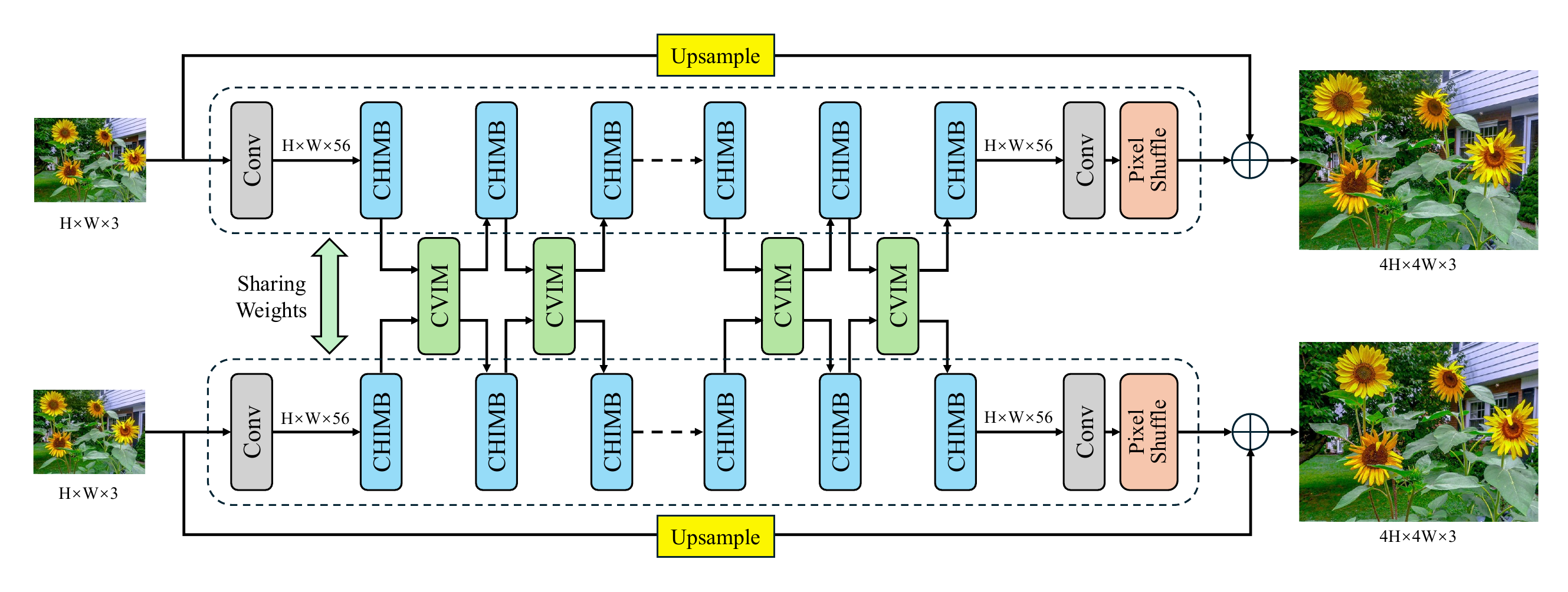}
        \caption{BUPTMM: The structure of the proposed Efficient CVHSSR.}
        \label{fig:BUPTMM}
    \end{figure*}

    This team developed a Cross-View Hierarchy Network for Stereo Image Super-Resolution (CVHSSR) \cite{Zou2023Cross} by leveraging the complementary information between different viewpoints (Fig.~\ref{fig:BUPTMM}). CVHSSR consits of two modules: the Cross-Hierarchy Information Mining Block (CHIMB) and the Cross-View Interaction Module (CVIM). CHIMB is developed to simulate and recover intra-view information across different levels, employing large-kernel convolutional attention and channel attention mechanisms. Meanwhile, CVIM utilizes a cross-view attention mechanism to effectively consolidate similar information from different views. These two modules facilitate CVHSSR to better aggregate cross-view information for higher performance.
    \par
    In order to improve the SR performance of the model more efficiently with a limited number of parameters and MACs, the number of channels and the number of the CHIMB and CVIM modules are tuned. The final number of parameters is 999.54 K. 
    \\
    \noindent \textbf{Training Settings.}
    The training of the proposed model consists three phases. In the first phase, only 700 stereo image pairs were used as the training set. The proposed model was  trained for 200K iterations with an MSE loss and a frequency Charbonnier loss. The Lion optimizer was employed in this phase. The batch size was set to 18 and the patch size was set to $64\times64$. The learning rate was initialized as $1 \times 10^{-4}$ and updated using a cosine annealing strategy. The minimum learning rate was set to $1 \times 10^{-8}$. Data augmentation was performed through RGB channel shuffing and horizontal/vertical flipping. In the second phase, the resultant model was used for initialization and all 800 stereo image pairs were included for training. The model was further trained for 100K iterations, with the same settings as the first phase. In the third phase, the resultant model is further trained for 100K iterations. The batch size was set to 16 and the learning rate was updated to $2 \times 10^{-6}$ for further training.
   
    \subsection{CV\_IITRPR - Track 1, 2}

    \begin{figure}[t]
        \centering
        \includegraphics[width=1\linewidth]{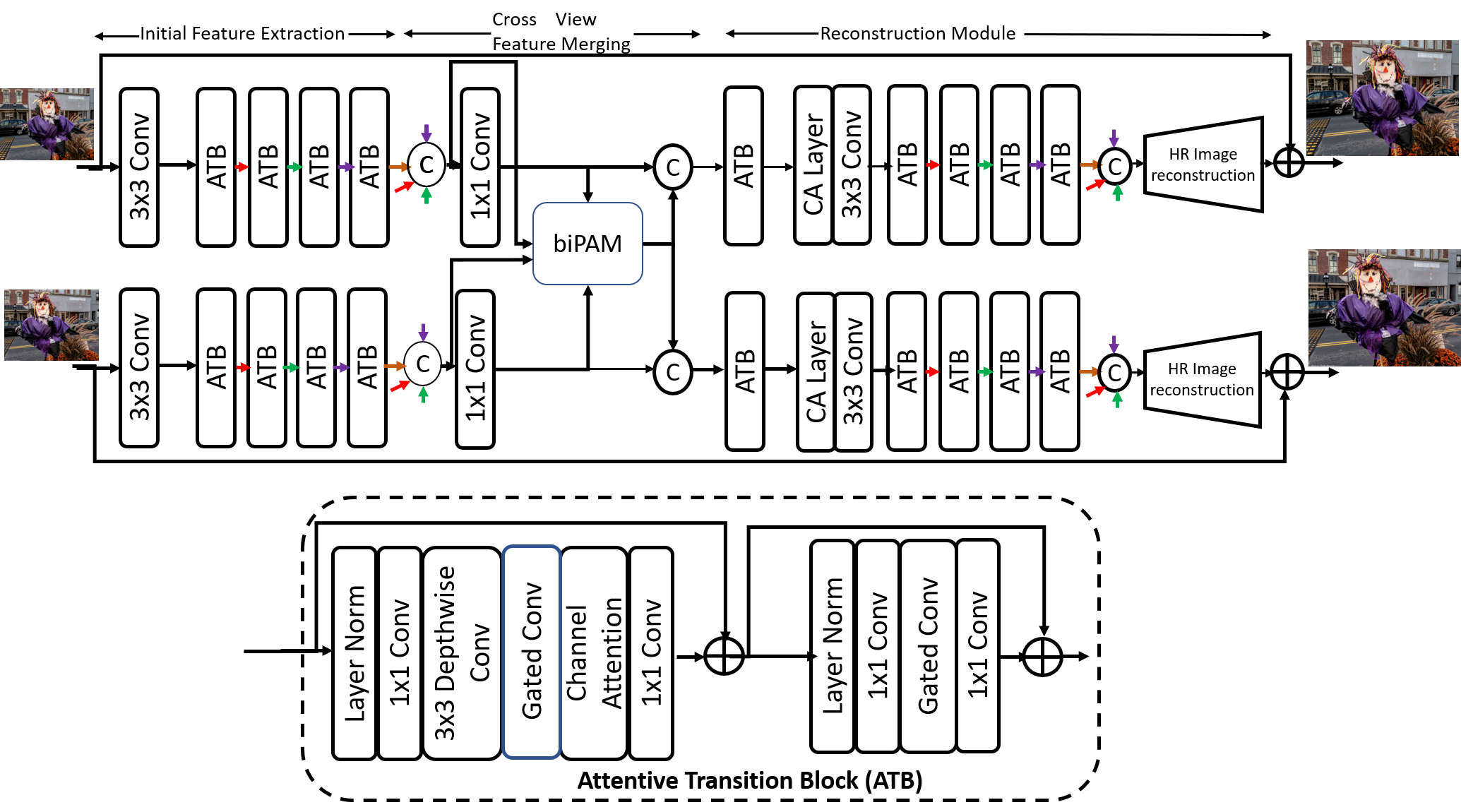}
        \caption{CV\_IITRPR: The structure of the proposed network.}
        \label{CVIITRPR}
    \end{figure}

    This team proposed a stereo image SR network with a two-branch structure. As shown in Fig.~\ref{CVIITRPR}, the proposed network consists of three stages, including initial feature extraction, cross-view feature merging, and reconstruction. Within the initial feature extraction module, attentive transition blocks are developed for better exploitation of features at different channels. After that, a bidirectional parallax attention module \cite{Wang2021Symmetric} is employed to aggregate features from both views. Finally, intra-view and inter-view features are collected to produce the super-resolved results.  
    
    \subsection{webbzhou - Track 1, 2}
\begin{figure}[t]
        \centering
        \includegraphics[width=1\linewidth]{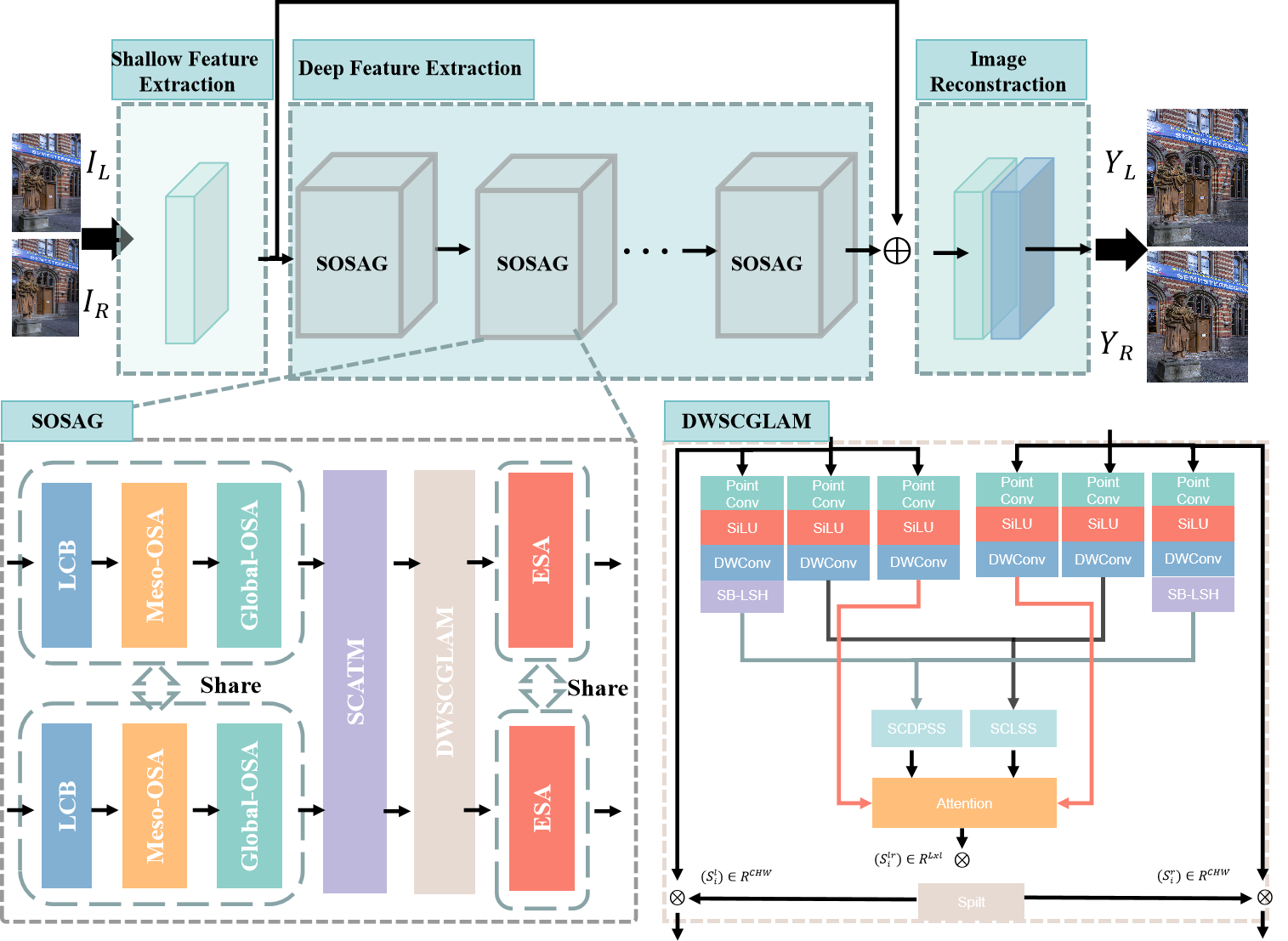}
        \caption{webbzhou: The structure of the proposed SOAN model.}
        \label{webbzhou-1}
    \end{figure}

    This team proposed a {S}tereo {O}mnidirectional {A}ggregation {N}etwork (SOAN) to efficiently super-resolve the stereo images, as illustrated in Fig.~\ref{webbzhou-1}. SOAN consists of three parts: shallow feature extraction, deep feature extraction, and image reconstruction modules. The shallow feature extraction module comprises a $3\times3$ convolution, while the image reconstruction module consists of convolutional layers and pixel shuffle layers. The deep feature extraction part consists of MOmni-Scale Stereo Aggregation Groups (OSSAG), with each OSSAG comprising LCB \cite{wang2023omni}, Meso-OSA~\cite{wang2023omni}, Global-OSA~\cite{wang2023omni}, SCATM~\cite{zhou2023stereo}, DWSCGLAM, and ESA~\cite{liu2020residual}. Specifically, LCB, Meso-OSA, and Global-OSA are used to aggregate local information at different scales. In addition, SCATM~\cite{zhou2023stereo} is adopted to aggregate features from two views. %Afterwards, DWSCGLAM is mainly based on the similarity intra and inter images to globally search for and compensate for the details lost in low-resolution images, the composition of which can be seen in Fig.~\ref{webbzhou-1}.
    
    \textbf{Track 1 Training Settings.} Considering that rotational augmentation would disrupt the epipolar constraints, it cannot be employed for data augmentation. To address this issue and enhance the robustness of the model, this team divided the training into two stages:
    \begin{itemize}
        \item At the first stage, the proposed model was trained on the Flickr1024 dataset for single-image SR. The SCATM and DWSCGLAM modules are excluded at this stage. Random rotation and flipping are included for data augmentation. L1 loss was utilized for training with a batch size of 32. Cosine annealing schedule with 500,000 iterations was employed to update the learning rate.
        
        \item At the second stage, the resultant model was further fine-tuned for stereo image SR. The SCATM and DWSCGLAM modules were recovered and the batch size was set to 64. %As mentioned earlier, rotational augmentation is not included in this stage. Additionally, 
        RGB channel shuffling was introduced for data augmentation. Cosine annealing schedule with 500,000 iterations was employed to update the learning rate. Mean squared error (MSE) loss is used for training.
    
    \end{itemize}
     
     \textbf{Track 2 Training Settings.}
     For track 2, the model trained in track 1 was used for initialization and then further fined-tuned on the training set of track 2.

    \subsection{Qi5 - Track 1}
\begin{figure}[t]
        \centering
        \includegraphics[width=1\linewidth]{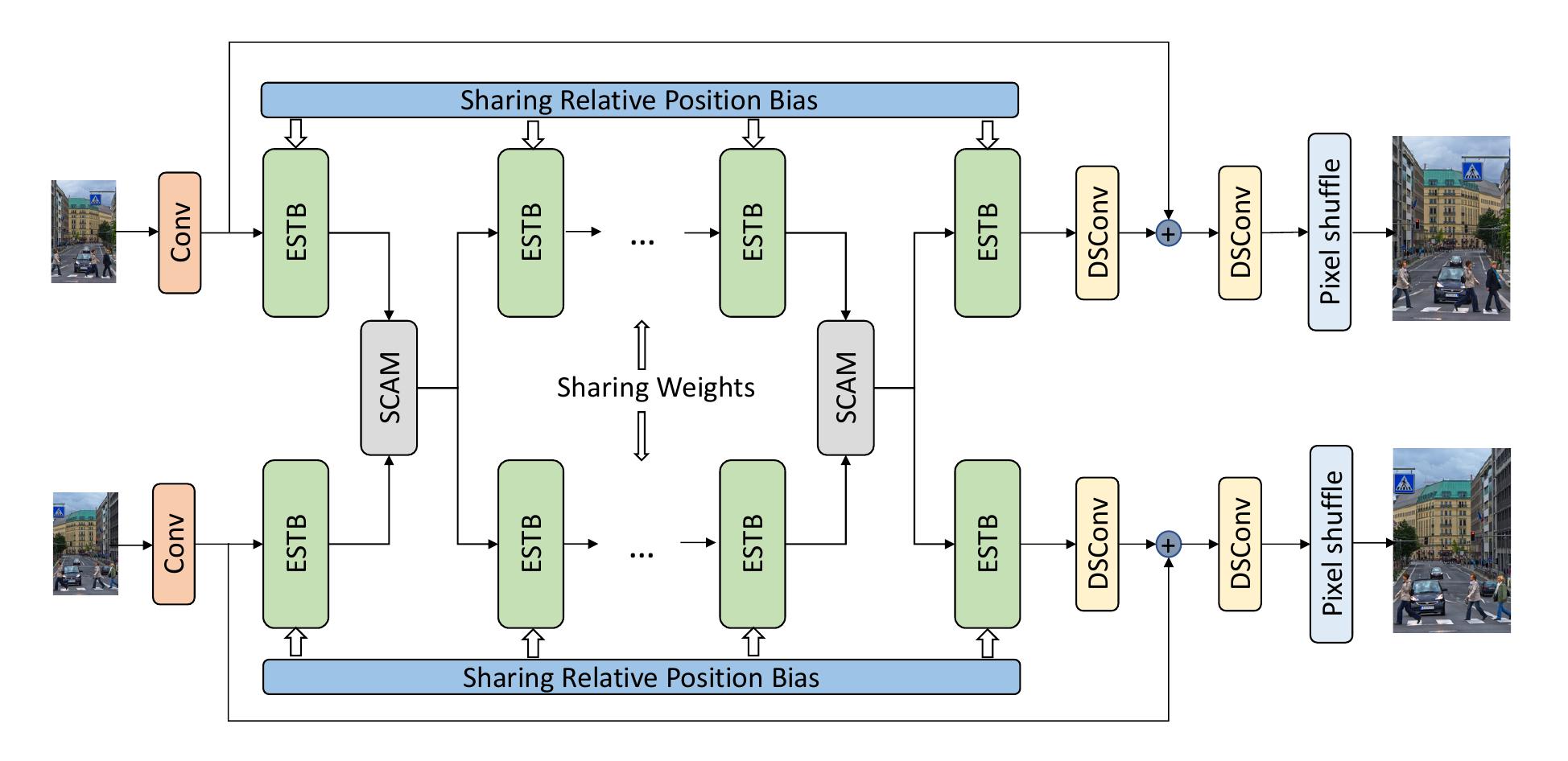}
        \caption{Qi5: The structure of the proposed efficient Swin Transformer network.}
        \label{Qi5-1}
    \end{figure}

     This team proposed an efficient Swin Transformer network for stereo image SR. First, they replaced all the vanilla convolutions in the deep feature extraction and image reconstruction stages with depth-wise convolutions. Second, they shared the relative position encoding parameters across all window attention modules, which allows the network to adopt larger window sizes while reducing parameters. Third, they removed the masking mechanism in shifted window multi-head self-attention (SW-MSA). In addition, the stereo cross-attention modules (SCAM) were employed to aggregate information from both views. The architecture is shown as Fig.~\ref{Qi5-1}
     
    \textbf{Training Settings.} During the training phase, the model was first trained with an L1 loss and then fine-tuned with an L2 loss. The Adam optimizer was utilized for optimization. The batch size was 16 and the patch size was $64 \times  64$. The learning rate was initiated as $\rm 2\times10^{-4}$ and halved at  150K, 250K, 350K, 450K, and 500K iterations.

    \begin{figure*}[htbp]
    % \begin{adjustwidth}{-\extralength}{0cm}
        \centering
        \includegraphics[width=\textwidth]{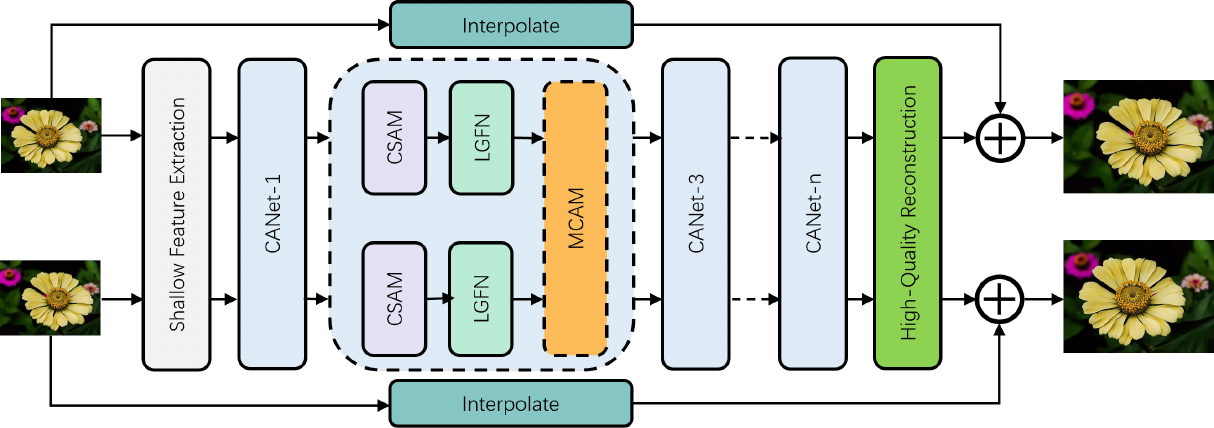}
        \includegraphics[width=\textwidth]{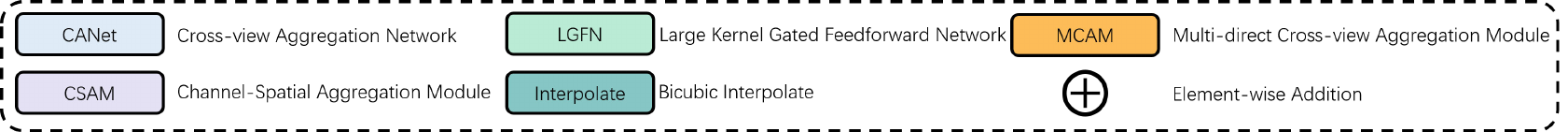}
        % \end{adjustwidth}
    \caption{WITAILab: The structure of the proposed CANSSR.}
    \label{WITALab-1}
\end{figure*}

    \begin{figure}[htbp]
        % \begin{adjustwidth}{-\extralength}{0cm}
            \centering
            \includegraphics[scale=0.54]{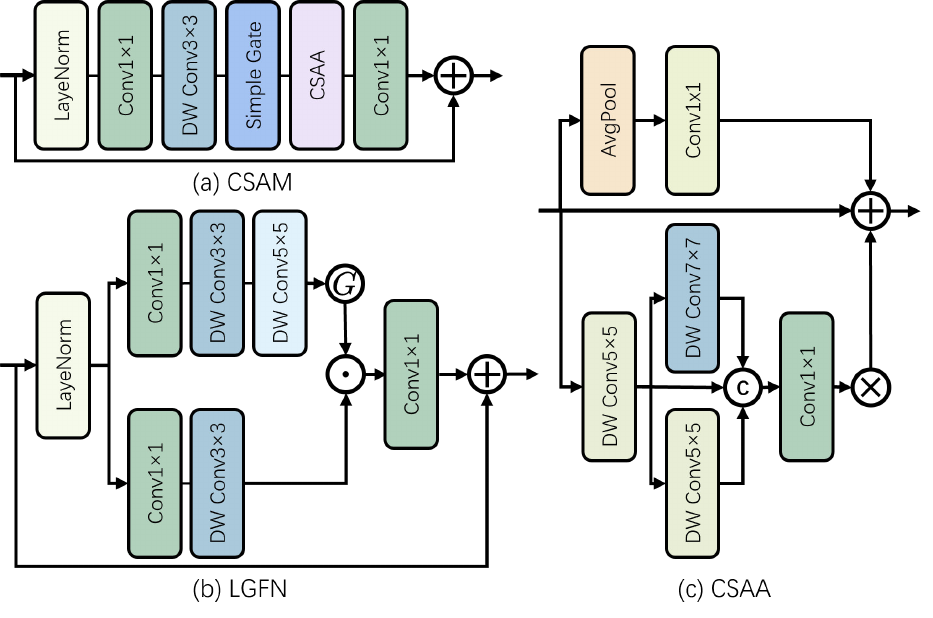}
            \includegraphics[scale=0.57]{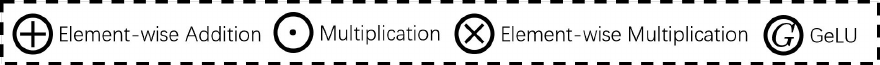}
            % \end{adjustwidth}
        \caption{WITAILab: The structure of the proposed modules. (a) Channel-Spatial Aggregation Module (b) Large Kernel Gated Feedforward Network (c) Channel-Spatial Aggregation Attention}
        \label{WITALab-2}
    \end{figure}

    % \end{figure}

    \subsection{WITAILab - Track 1}
    %There remains an underexplored and inadequately balanced aspect concerning intra-view and cross-view similarity information, resulting in unsatisfactory reconstruction outcomes. Given the homogeneous observation characteristics of the current scene by binocular cameras, intra-view and cross-view similarities exist for feature representation.

\begin{figure}[htbp]
        % \begin{adjustwidth}{-\extralength}{0cm}
            \centering
            \includegraphics[scale=0.53]{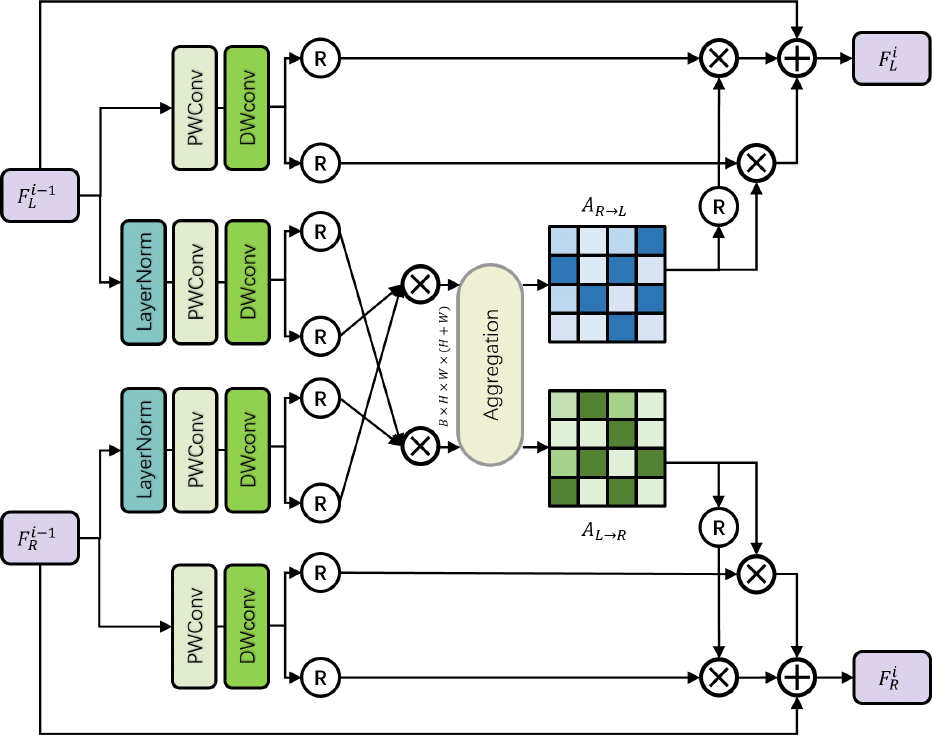}
            % \end{adjustwidth}
        \caption{WITAILab: The structure of the proposed multi-direction cross-view aggregation module.}
        \label{WITALab-3}
    \end{figure}

    This team proposed a novel method named cross-view aggregation network for stereo image super-resolution(CANSSR) which exploits multi-directional parallax attention to capture both horizontal and vertical stereo correspondences while enhancing long-range dependence.
	Specifically, they proposed a multi-directional cross-view aggregation module(MCAM, as shown in Fig.~\ref{WITALab-3}) to aggregate the horizontal and vertical stereo correspondences to obtain more reliable complementary information from cross-view. Furthermore, to effectively aggregate the high-frequency detailed information for intra-view, we propose a channel-spatial aggregation module(CSAM, as shown in Fig.~\ref{WITALab-2}(a)) to capture long-range dependencies. Finally, we introduce a large-kernel convolution feed-forward network(LGFN, as shown in Fig.~\ref{WITALab-2}(b)) to aggregate richer spatial texture information, and a non-linear free activation function is introduced to enhance the non-linear representation. The overall architecture is shown as Fig.\ref{WITALab-1}.

    % This team improved the performance of stereo image SR by exhaustively exploring similar representations from multiple views in stereo pairs and introducing a novel cross-view aggregation network (CANSSR). Specifically, they proposed a bi-directional cross-view aggregation module (BCAM, as shown in Fig.~\ref{WITALab-2}(c)) and a global-local aggregation module (GLAM, as shown in Fig.~\ref{WITALab-1} and Fig.~\ref{WITALab-2}(a)) to explore and aggregate intra-view and cross-view information. To fully aggregate the global information within one view, a large convolution kernel is employed to capture contexts in larger receptive fields. The overall architecture is shown as Fig.\ref{WITALab-3}.

    \textbf{Training Settings.} All the models were optimized using AdamW with $\beta_1$ = 0.9 and $\beta_2$ = 0.999. The learning rate was initialized as $1\times10^{-3}$ and decayed to $1\times10^{-7}$ using the cosine annealing strategy. Batch size was set to 16. The proposed model was trained for 400,000 iterations using an MSE loss on two NVIDIA RTX 4090 GPUs.

    \subsection{Giantpandacv - Track 1, 2}
\begin{figure}[t]
        \centering
        \includegraphics[width=1\linewidth]{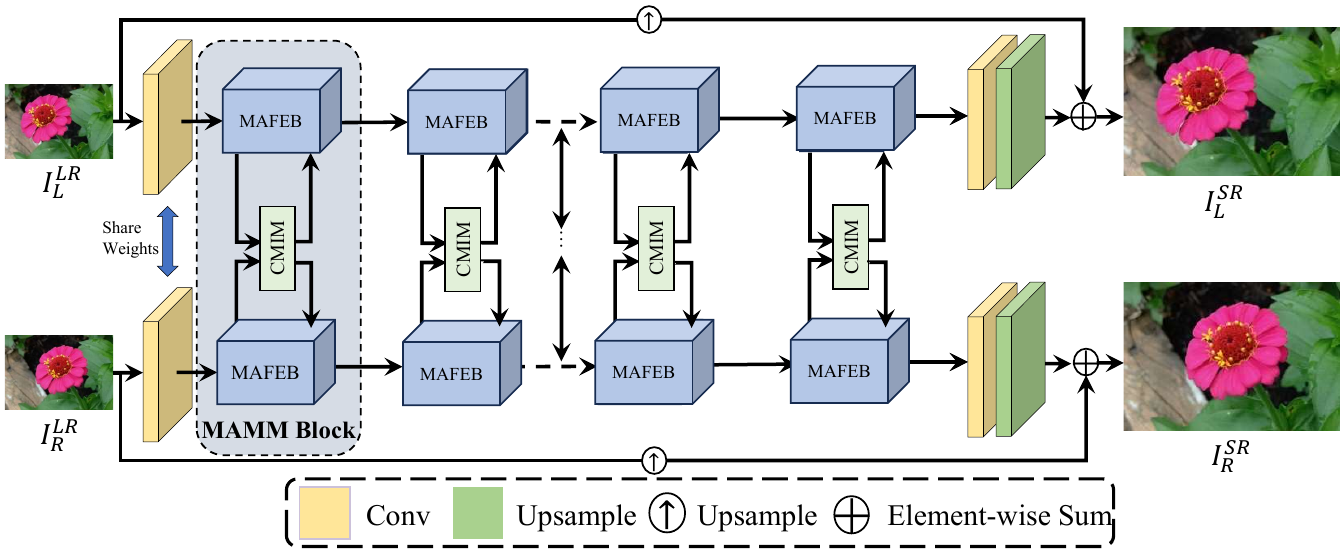}
        \caption{Giantpandacv: An overview of the efficient multi-level information extraction network for stereo image super-resolution (MIESSR).}
        \label{GPCV-1}
    \end{figure}

\begin{figure}[t]
        \centering
        \includegraphics[width=1\linewidth]{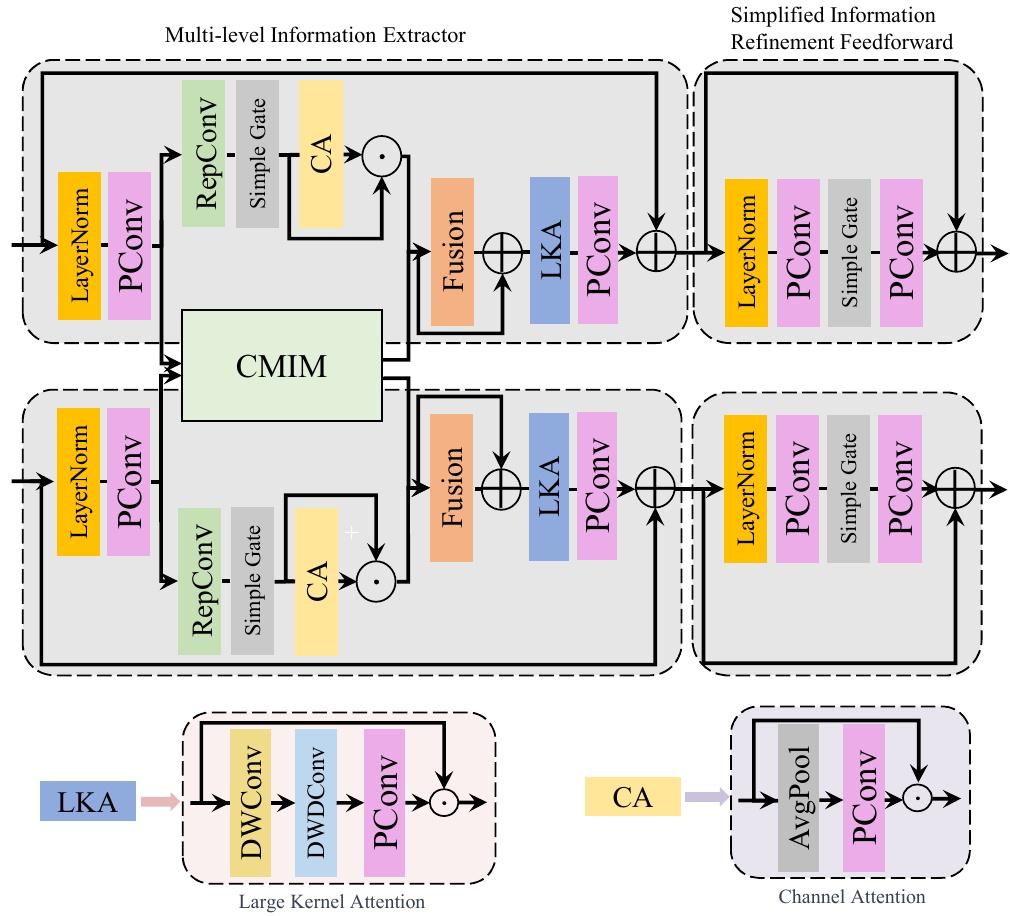}
        \caption{Giantpandacv: The structure of the proposed mixed attention feature extraction block (MAFEB).}
        \label{GPCV-2}
    \end{figure}

\begin{figure}[t]
        \centering
        \includegraphics[width=1\linewidth]{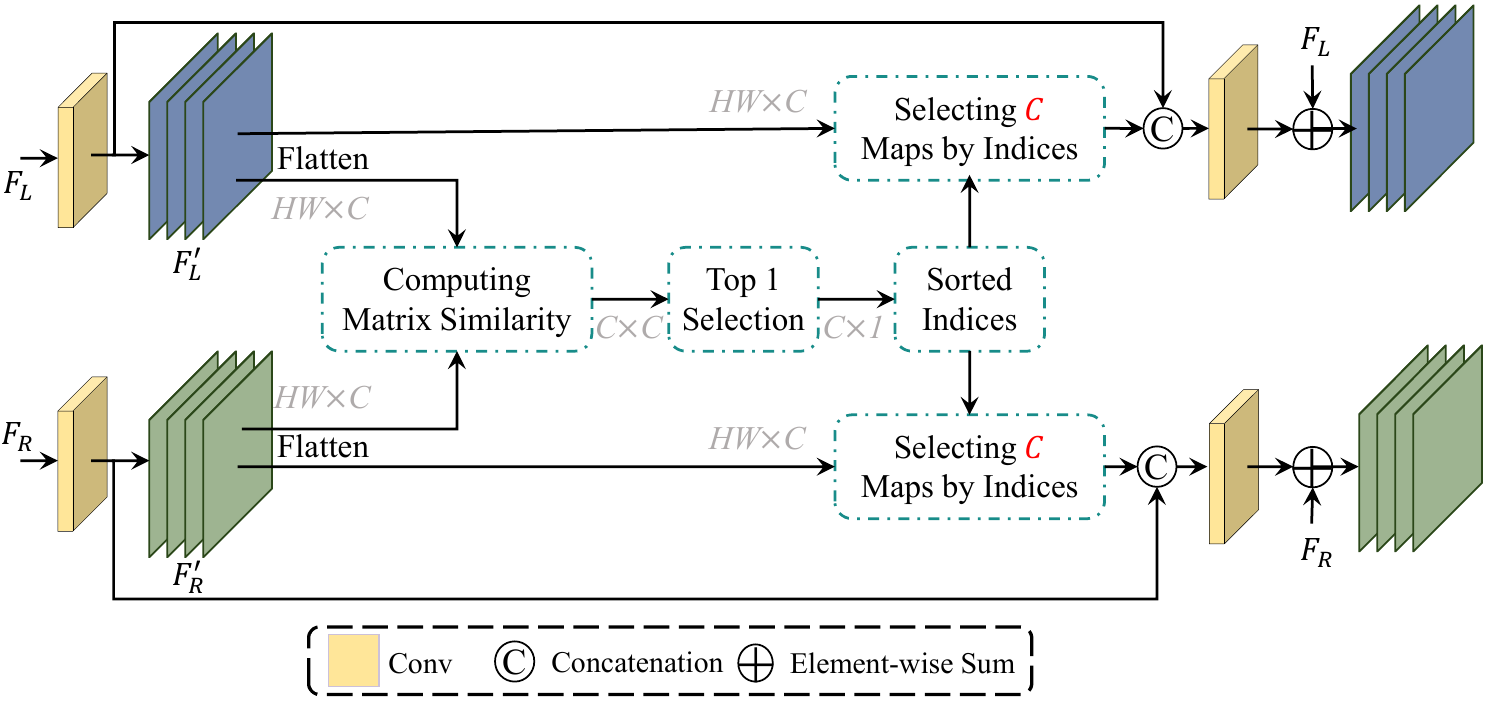}
        \caption{Giantpandacv: The structure of the proposed cross-view matching information modules (CMIM).}
        \label{GPCV-3}
    \end{figure}

    This team proposed an efficient multi-level information extraction network for stereo image SR (MIESSR), as shown in Fig.~\ref{GPCV-1}. Specifically, MIESSR consists of mixed attention feature extraction blocks (MAFEB, as shown in Fig.~\ref{GPCV-2}) and correlation matching information modules (CMIM, as shown in Fig.~\ref{GPCV-3}). MAFEBlocks aims to extract multi-scale view features and efficiently interact with CMIMs. CMIMs are mainly used to extract cross-view differential features and fully utilize the complementary information of stereo images. %This structural configuration can efficiently mine features inside the view, improving the efficiency of cross-view information sharing. Hence, reconstruct image details and textures more accurately.     More details of MAFEB and CMIM are described as follows.

    \textbf{Mixed Attention Feature Extraction Block.} %Stereo images contain multi-level information at global, local, and cross-view ranges, which could be utilized to better restore high-frequency details. Firstly, convolutional neural networks (CNNs) with small kernels can be utilized to acquire local information. Secondly, %the global range includes the resemblance of texture and structure, as well as consistency of content and semantics across various locations in the image. Effective whole-perception is vital for the purpose to acquire global information. Based on these ideas, they proposed a mixed attention feature extraction block is developed to efficiently capture both local and global information in the image, as shown in Fig.~\ref{GPCV-2}. 
    The MAFEB consists of two parts: (1) The multi-level information extractor (MIE), and (2) The simplified information refinement feedforward network (SIRFFN). In addition to introducing channel attention and large kernel convolution attention, MIE employs reparameterized convolution to further improve the model capacity and flexibility. %Thus, the block can adjust to structural features of various scales and shapes. 
    Besides, %they only used partial view features that interact with the embedded CMIM by utilizing branch structures to further improve efficiency. They use 
    SIRFFN is adopted to reduce structural redundancy. 

    \textbf{Cross-View Matching Module.} 
    Parallax attention has been widely used in previous stereo image SR works to aggregate information from both left and right views. Nevertheless, parallax calculation introduces considerable computational cost. In addition, erroneous textures and artifacts can be introduced due to inaccurate correspondence. To remedy this, this team developed a cross-view matching information module (CMIM) for cross-view information interaction, as shown in Fig.~\ref{GPCV-3}. %It is based on Correlation Matching Transformation, which computes the matrix similarity of the cross-views and selects feature maps by sorted indices. 
    Specifically, CMIM first computes the similarity along the channel dimension between left and right view feature $F_L, F_R$ to generate similarity matrix $M \in \mathcal{R}^{C \times C}$. Then, features with the highest similaries $F_L^{Select}, F_R^{Select}$ were selected from $F_L, F_R$ and concatenated with $F_L, F_R$ for subsequent feature aggregation.  Note that, the channel and block number of the NAFBlock were set to 56 and 24 in this method, which meets the requirements of the model size and computational complexity.

    \textbf{Training Settings.} The numbers of MAFEB blocks and feature channels are set to 64 and 20, respectively. The whole network was trained using an MSE loss and a Frequency Charbonnier loss, and optimized using the Lion method \cite{chen2024symbolic} with $\beta_1$=0.9, $\beta_2$=0.999, and a batch size of 64. The proposed CVHSSR was implemented using PyTorch on a PC with four NVidia A100 GPUs. The learning rate was initially set to $5\times10^{-4}$ and decayed using a cosine annealing strategy. The proposed model was trained for 200,000 iterations. %Finally, They fine-tuned the network using both MSE loss and frequency Charbonnier loss for 200,000 iterations with a learning rate $5\times10^{-5}$.

    \subsection{JNU\_620 - Track 1, 2}
\begin{figure*}[t]
        \centering
        \includegraphics[width=1\linewidth]{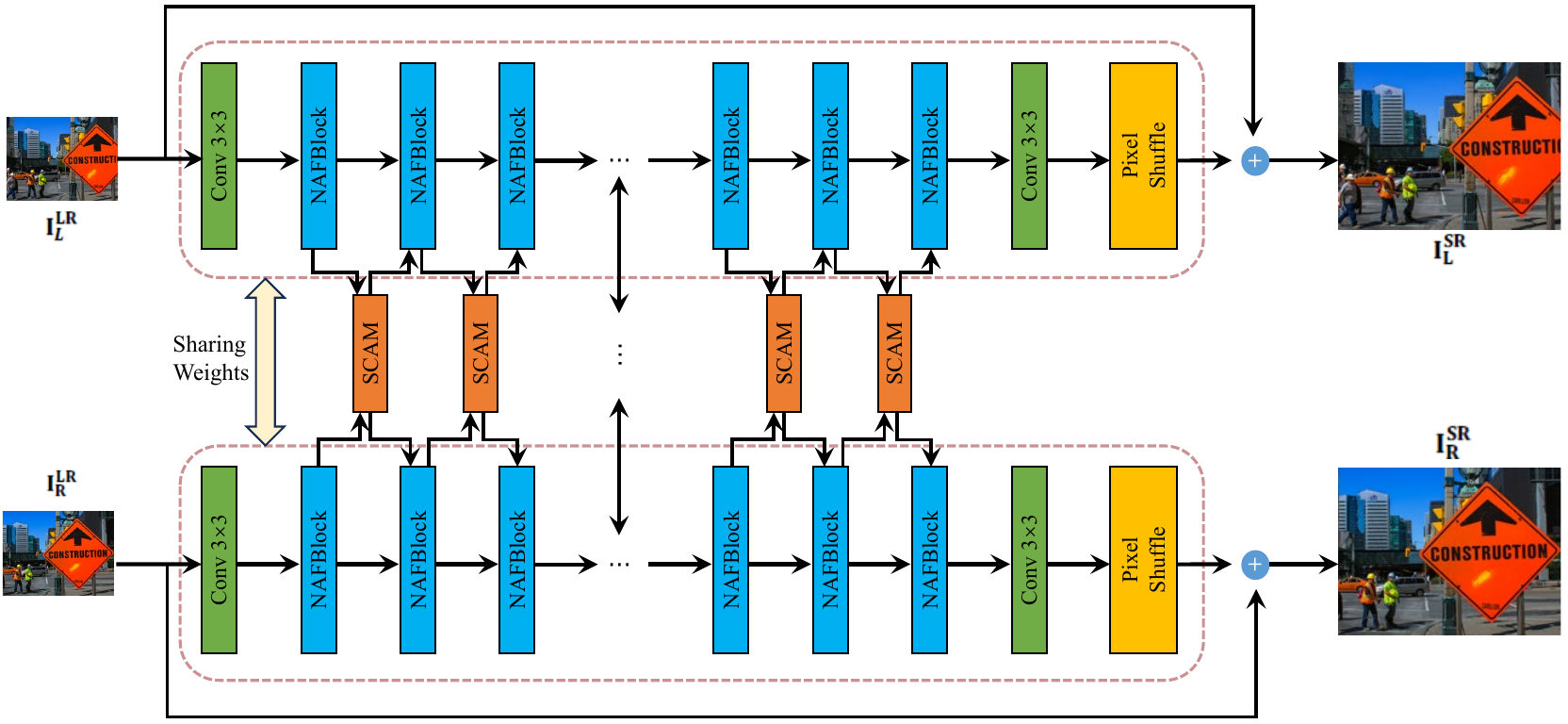}
        \caption{JNU\_620: The structure of the proposed modified NAFSSR.}
        \label{JNU620-1}
    \end{figure*}

    This team contructed their solution based on NAFSSR~\cite{Chu2022NAFSSR}, as shown in Fig.~\ref{JNU620-1}. For different tracks, different losses were proposed for training.
    
    \textbf{Track 1 Training Settings.}
    A back-projection (BP) loss~\cite{li2022best} was introduced for optimization, which enforces that the downscaled super-resolved images match the original LR observations:
    \begin{equation}
    L_{BP} = {\left| \left|  S({I}_{L}^{SR},s)-I_{L}^{LR}   \right| \right|}_1 + {\left| \left|  S({I}_{R}^{SR},s)-I_{R}^{LR}    \right| \right|}_1,
    \end{equation}
    where $S$ denotes the bicubic downsampling operation and $s$ represents the downscale factor (\textit{i,e.}, 4). The overall loss function is formulated as:
    \begin{equation}
    L_{total} = L_1 + \lambda L_{BP},
    \end{equation}
    where $\lambda$ is set to $0.1$.
    
    \textbf{Track 2 Training Settings.}
    Since the realistic degradation is more complex than the bicubic degradation, only an L1 loss was used for training. During the training phase, the network was trained for $4 \times$ SR with a batch size of 24 and a patch size of $30 \times 90$. The AdamW method was employed for optimization with $\beta_1 = 0.9$ and $\beta_2 = 0.9$. The initial learning rate was set to $3\times10^{-3}$, and decreased to $1 \times 10^{-7}$ with a cosine annealing strategy. Random horizontal/vertical flipping and RGB channel shuffling were adopted for data augmentation. %In the last stage, an MSE loss was adopted for fine-tuning.

    \begin{figure}[t]
        \centering
        \includegraphics[width=1\linewidth]{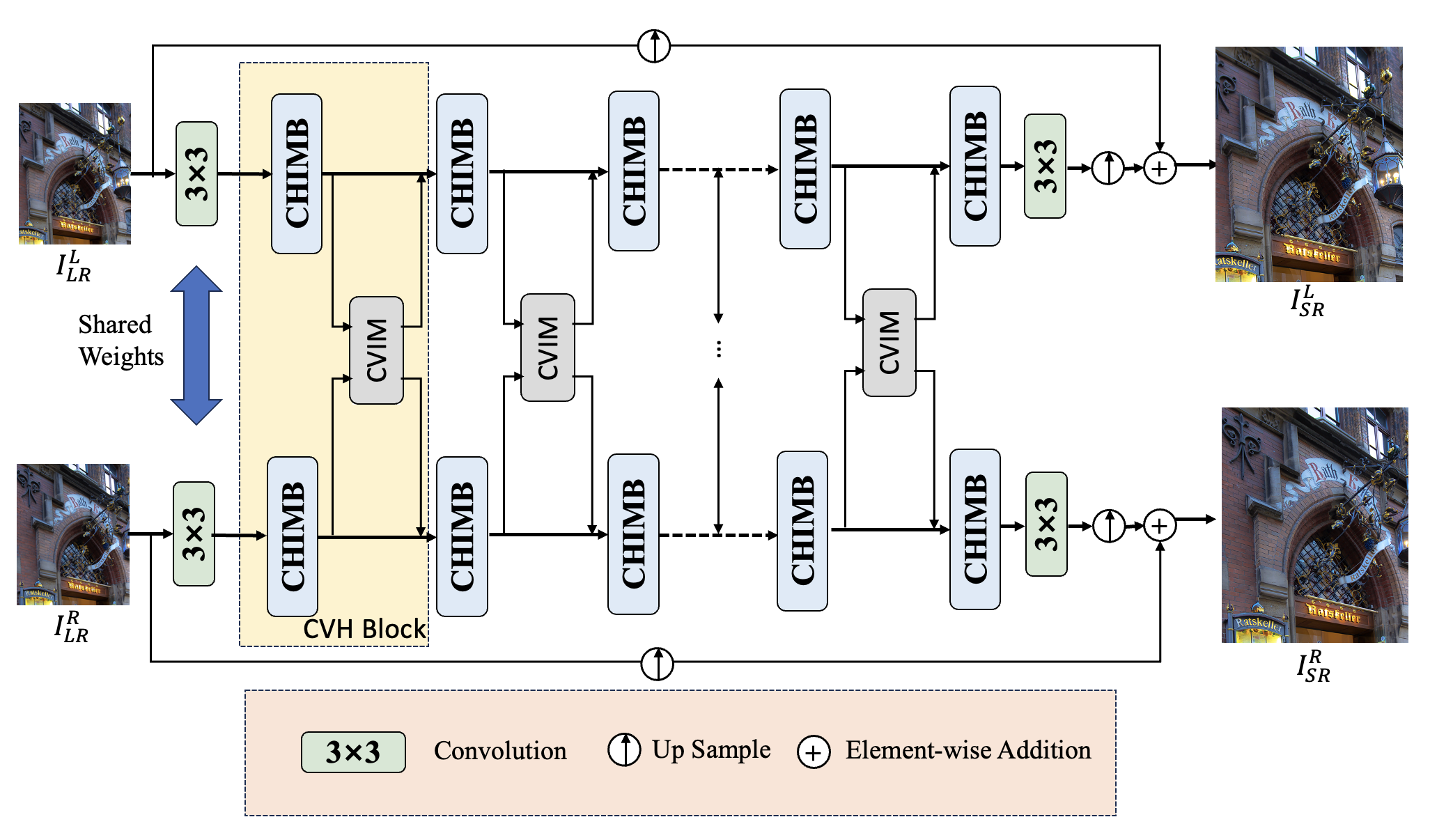}
        \caption{GoodGame: The structure of the proposed CVHSSRPlus model. CVIM and CHIMB in addition to the Information Refinement Feedforward module is the same as CVHSSR~\cite{Zou2023Cross}}
        \label{GoodGame-1}
    \end{figure}

    \begin{figure}[t]
        \centering
        \includegraphics[width=1\linewidth]{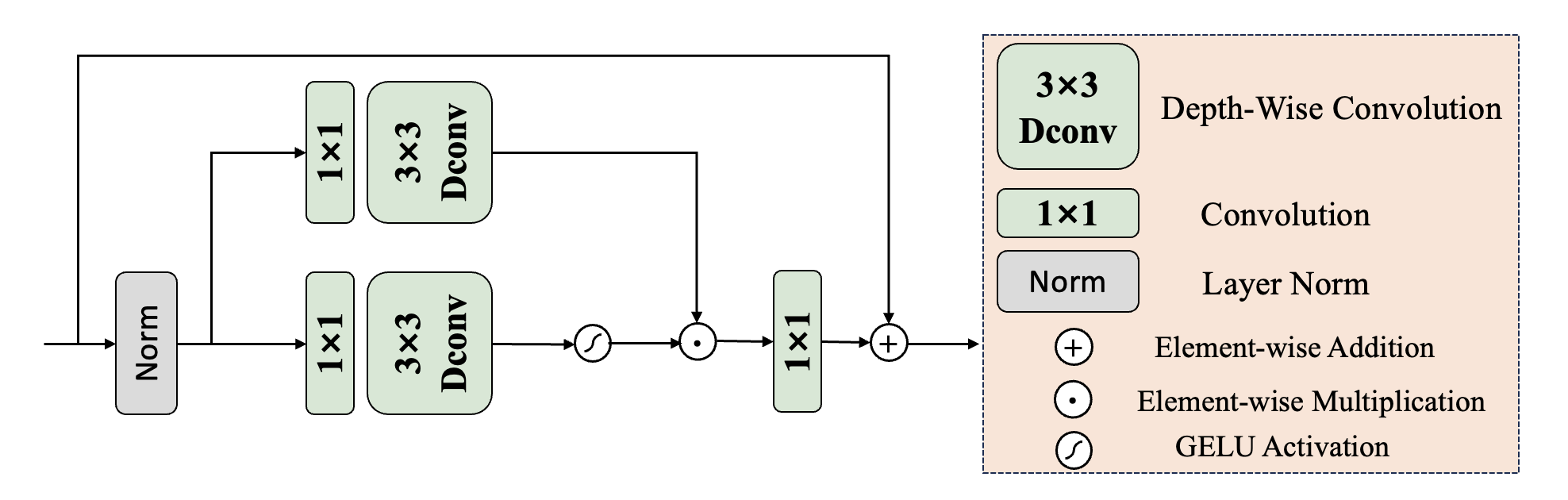}
        \caption{GoodGame: The structure of the proposed module that is used to replace Information Refinement Feedforward for feature extraction.}
        \label{GoodGame-2}
    \end{figure}
    
     \subsection{GoodGame - Track 1, 2}
     This team used the CVHSSR~\cite{Zou2023Cross} model (Figure~\ref{GoodGame-1}) as the baseline. The proposed method pays special attention to the complementary information between left and right views in stereo image pair, as well as the importance of the information within the respective views. In CVHSSR, CHIMB uses channel attention and large-kernel convolutions to extract global and local features within each view. CVIM aims to fuse features from different views through a cross-view attention mechanism. To further boost CVHSSR, the Information Refinement Feedforward module in CHIMB was replaced with the structure shown in Fig.~\ref{GoodGame-2}.

     \textbf{Training Settings.}
    The number of CVH Blocks was set to 17, and the total parameter amount of the model was approximately 985K. During the training phase, images were cropped to $128\times128$ patches with horizontal/vertical flipping being employed for data augmentation.

    \begin{figure*}[t]
        \centering
        \includegraphics[width=1\linewidth]{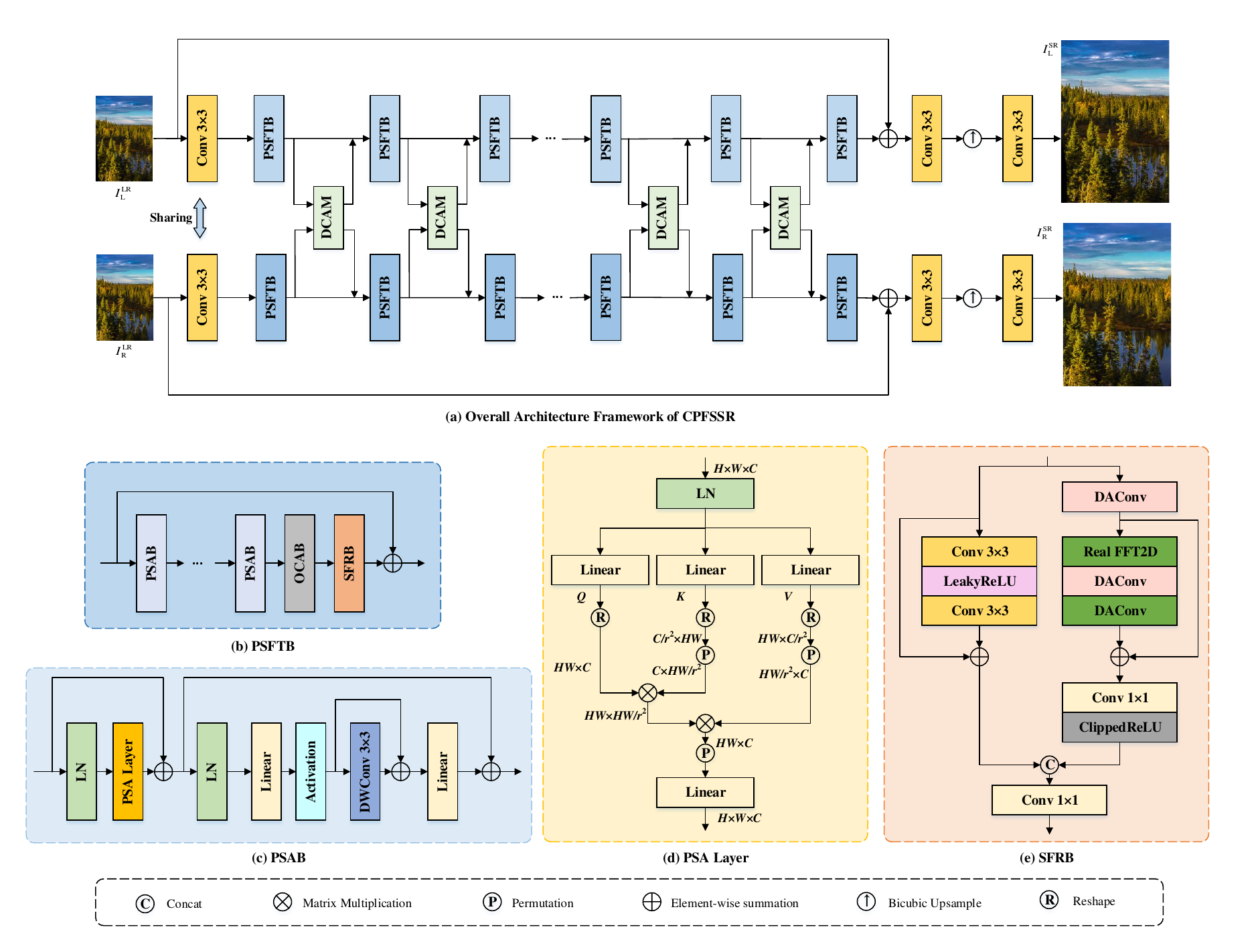}
        \caption{Fly\_Flag: The structure of the proposed CPFSSR.}
        \label{FlyFlag}
    \end{figure*}
    
    \subsection{Fly\_Flag - Track 1, 2}
   This team developed a model termed combined substitution self-attention and fast fourier convolution for stereo image super-resolution (CPFSSR). The framework of the proposed CPFSSR is shown in Figure~\ref{FlyFlag}. Specifically, they constructed permuted Swin Fourier Transformer block (PSFTB, as shown in Figure~\ref{FlyFlag} (b)) based on permutation self-attention (PSA, as shown in Figure~\ref{FlyFlag} (d)) layer, better captured global features and local features with limited window size size by cascading Permuted Self-Attention Blocks (PSABs, as shown in Figure~\ref{FlyFlag} (c)), and built a spatial frequency reinforcement block (SFRB, as shown in Figure~\ref{FlyFlag} (e)) based on the fast Fourier convolution block to improve the extraction of frequency domain information. In addition, they designed a deep cross-attention module (DCAM) to model the depth of hierarchical information before interaction, which facilitates the realization of information interaction between different views.
    
    % Inspired by the observation that permutation self-attention (PSA)~\cite{zhou2023srformer} strikes a balance between self-attentive channels and spatial information, this team modified the hybrid attention block (HAB) in the RHAG~\cite{chen2023activating} through the permutation self-attention (PSA) mechanism to better utilize the spatial information within a limited window. To further capture and fuse the global and local features, this team proposed a spatial frequency reinforcement block (SFRB). The SFRB consists of a frequency branch and a regular spatial branch. Specifically, fast Fourier convolutions are employed in the frequency branch while vanilla convolutions are used in the residual branch. Further, they replace the convolutional layer of the RHAG module with SFRB and use the permuted Swin Fourier Transformer block (PSFTB) for feature extraction. Besides, this team proposed a cross-hierarchy interaction module (CHIM) to capture global and local similarity relations through channel attention and large kernel convolutional attention.

    \textbf{Training Settings.} A pixel loss and a frequency loss are used for training:
    \begin{flalign}
        &\mathcal{L}=\mathcal{L}_{\mathrm{Charbonnier}}+\lambda \mathcal{L}_{\mathrm{FFT}},	  &\\
        &{\mathcal{L}_{\mathrm{Charbonnier}}} = \frac{1}{N}\sum\limits_{i = 1}^N {\sqrt {{{\left\|  { {I_\mathrm{L,R}^{\mathrm{HR}}} - {I_\mathrm{L,R}^{\mathrm{SR}}}} \right\|}^2} + { \varepsilon_1^2}} } ,	 &\\
        &{\mathcal{L}_{\mathrm{FFT}}} = \frac{1}{N}\sum\limits_{i = 1}^N {\sqrt {{{\left\| {\mathrm{FFT}\left( {I_\mathrm{L,R}^{\mathrm{HR}}} \right) - \mathrm{FFT}\left( {I_\mathrm{L,R}^{\mathrm{SR}}} \right)} \right\|}^2} + { \varepsilon_2^2}} } ,	  &
    \end{flalign}
    where $\varepsilon_1$ is set to $10^{-3}$, ${I_\mathrm{L,R}^{\mathrm{SR}}}$ represents the left and right SR image, and ${I_\mathrm{L,R}^{\mathrm{HR}}}$ represents the corresponding HR image. $\varepsilon_2$ is set to $10^{-3}$ and $\mathrm{FFT}(\cdot)$ represents the Fast Fourier Transform. $\lambda$ is a hyperparameter that controls the frequency Charbonnier loss function and set to 0.1.

    During training, the Adam method with $\beta_1=0.9$ and $\beta_2=0.99$ was employed for optimization. The learning rate was initialized to $2\times10^{-4}$ and reduced to $1\times10^{-7}$ using the cosine annealing strategy. %During training, each batch size of 16 samples was evenly distributed across four partitions running on four Nvidia A100 GPUs for 3 × $10^5$ iterations.
    Random cropping, random vertical and horizontal flipping, random horizontal shifting, and random RGB channel shuffling are adopted for data augmentation.

    \begin{figure}[t]
        \centering
        \includegraphics[width=1\linewidth]{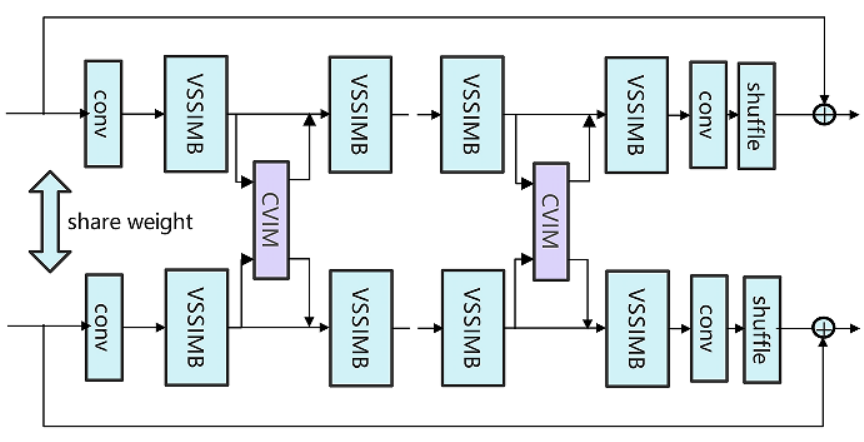}
        \caption{Mishka: The structure of the proposed VSSSR.}
        \label{Mishka1}
    \end{figure}

    \begin{figure*}[t]
        \setcounter{figure}{21}
        \centering
        \includegraphics[width=0.9\linewidth]{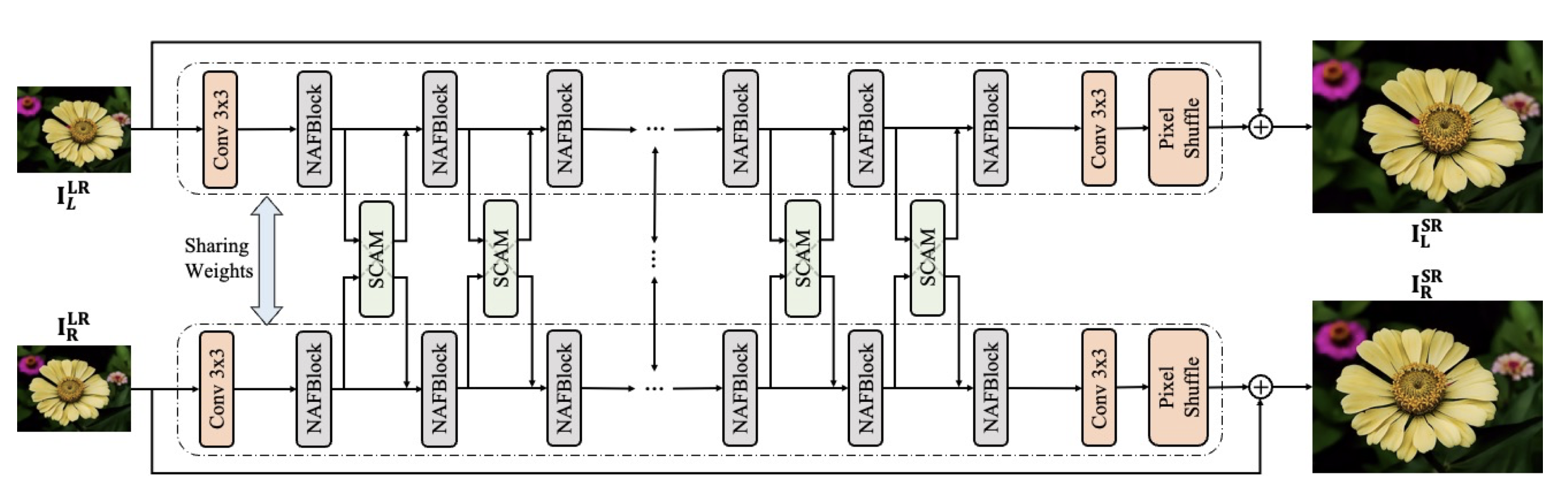}
        \caption{DVision: The structure of the proposed modified NAFSSR. This team modifies NAFSSR by using ODConv layers and knowledge distillation.}
        \label{DVision}
    \end{figure*}
    
    \begin{figure}[t]
        \setcounter{figure}{19}
        \centering
        \includegraphics[width=1\linewidth]{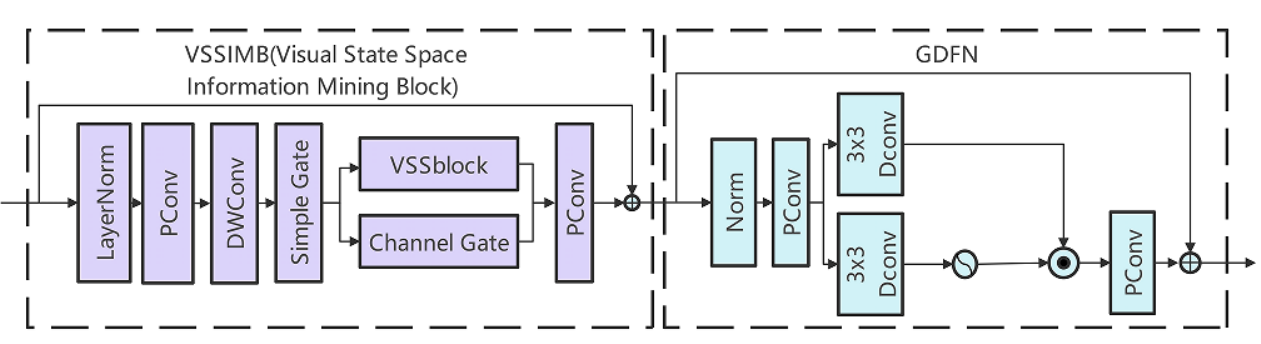}
        \caption{Mishka: The structure of the visual state Space Information Mining Block (VSSIMB).}
        \label{Mishka2}
    \end{figure}

    \subsection{Mishka - Track 1, 2}
    This team developed a VSSR network based on CVHSSR \cite{Zou2023Cross}, as illustrated in Fig.~\ref{Mishka1}). Specifically, the VSSBlock from VMamba \cite{liu2024vmamba} was adapted to extract image features in the spatial dimension. Concurrently, the ChannelGate \cite{Woo2018Cbam} module is utilized to obtain image feature in the channel dimension, forming the VSSIMB (Fig.~\ref{Mishka2}).
    Additionally, we incorporated the GDFN (Gated-Dconv Feed-Forward Network) module from Restormer \cite{Zamir2022Restormer} to refine the features extracted by VSSIMB.

    \textbf{Training Settings.}
    For track 1, the proposed network was trained using an MSE loss function and a frequency Charbonnier loss. The model was optimized using the Lion method with $\beta_1 = 0.9$, $\beta_2 = 0.9$, and a batch size of 4 per GPU. The initial learning rate was set to $1\times10^{-4}$ and updated using a cosine decay strategy. The model was first trained for 200,000 iterations and then fine-tuned for an additional 200,000 iterations using an MSE loss with the learning rate being set to $1\times10^{-5}$. Additionally,  stochastic depth strategy was employed for network regularization, with the drop path rate being set to 0.1. For track 2, the model obtained from track 1 was used for initialization. Then, the model was further fine-tuned using an MSE loss with the same settings. %The Lion optimization method continued to be used with the learning rate set to $1e-5$, maintaining the same cosine decay strategy and stochastic depth approach.
    During the training phase, horizontal flipping, vertical flipping, and RGB flipping were adopted for data augmentation.
    
    \begin{figure}
        \setcounter{figure}{20}
        \centering
        \includegraphics[width=1\linewidth]{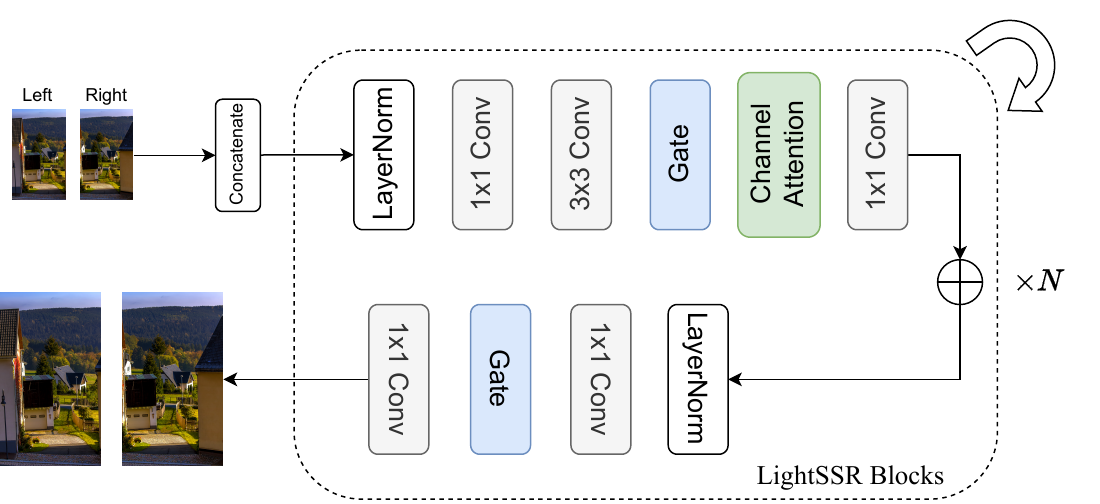}
        \caption{LightSSR: The structure of the proposed LightSSR block.}
        \label{LightSSR}
    \end{figure}
    
    \begin{figure*}[ht]
    \setcounter{figure}{24}
        \centering
        \includegraphics[width=0.8\textwidth]{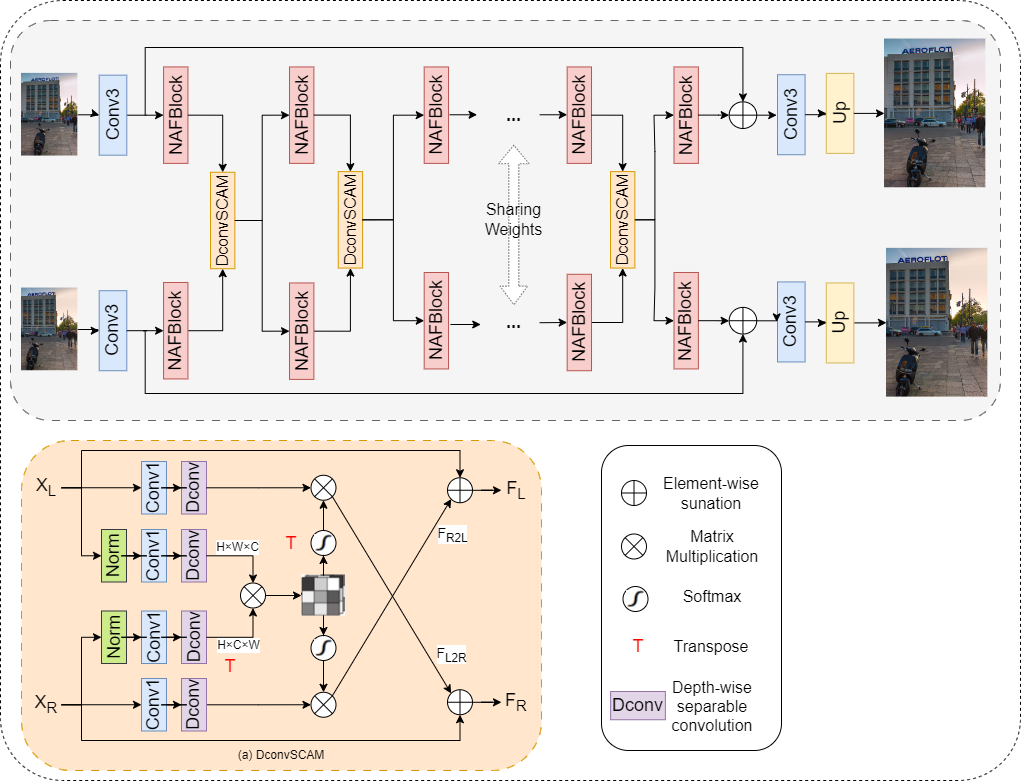}
        \caption{Liz620: The structure of the proposed DconvNAFSSR}
        \label{fig:DconvNAFSSR}
    \end{figure*}
    
    \subsection{LightSSR - Track 1}
    
    %The complexity of a Super-Resolution (SR) network can be decomposed into two primary components: inter-block complexity and intra-block complexity. Inter-block complexity arises from connections between feature maps of varying sizes, while the intra-block complexity pertains to the design choices within the block itself, such as the Multi-Dconv Head Transposed Attention Module introduced in \cite{zamir2022restormer} and the Swin Transformer Block in \cite{liu2021swin}. 
    Drawing inspiration from \cite{chen2022simple}, this team introduced an efficient and lightweight stereo super-resolution method named LightSSR. The complexity of a stereo image SR network can be decomposed into inter-block computational complexity and intra-block computational complexity. %Inter-block complexity arises from connections between feature maps of varying sizes, while the intra-block complexity pertains to the design choices within the block, such as the Multi-Dconv Head Transposed Attention Module introduced in \cite{Zamir2022Restormer} and the Swin Transformer Block in \cite{Liu2021Swin}. 
    To reduce inter-block computational complexity, they employed a single-stage UNet architecture. To achieve  savings in terms of intra-block computational complexity, they simplify the baseline from HINet \cite{Chen2021Hinet} by removing the connections between Gaussian Error Linear Units and Channel Attention to Gated Linear Unit. %Specifically, they streamline the model by either removing or replacing nonlinear activation functions, such as Sigmoid, ReLU, and GELU
    Figure~\ref{LightSSR} illustrates the basic blocks of our LightSSR model. %Each component in the model is trivial, such as the Layer Normalization, Convolution, GELU, and Channel Attention. However, the combination of these trivial components leads to a strong performance. 
    The hidden channel of the proposed network was to 48, and the number of blocks was set to 32. 
    %\paragraph{Gate Layer:} The gated linear units are simplified by replacing the non-linear activation function with an identity function. The Gate layer could be implemented by an element-wise multiplication: $Gate(\mathbf{X},\mathbf{Y})=\mathbf{X}\odot\mathbf{Y}$, where $\mathbf{X}$ and $\mathbf{Y}$ are feature maps of the same size. 

    %\paragraph{Channel Attention:} The channel attention is simplified by retaining the two most important roles, including aggregating global information and channel information interaction. The proposed channel attention layer is $CA(\mathbf{X}) = \mathbf{X} \ast \text{Conv}_{1\times1}(\text{Pool}(\mathbf{X}))$, where $\text{Conv}_{1\times1}$ is a $1\times1$ convolution layer, $\ast$ is a channel-wise product operation, and $\text{Pool}$ indicates the global average pooling operation which aggregates the spatial information into channels. 
    
    %\textbf{Training Settings.} The hidden channel of the proposed network was to 48, and the number of blocks was set to 32. %When dealing with stereo images, they fuse the left and right images through concatenation before feeding them into the network as input. Finally, they use a neural degradation algorithm \cite{luo2024and} to achieve data augmentation during training. 

    \begin{figure}
        \setcounter{figure}{22}
		\centering
		\includegraphics[width=1\linewidth]{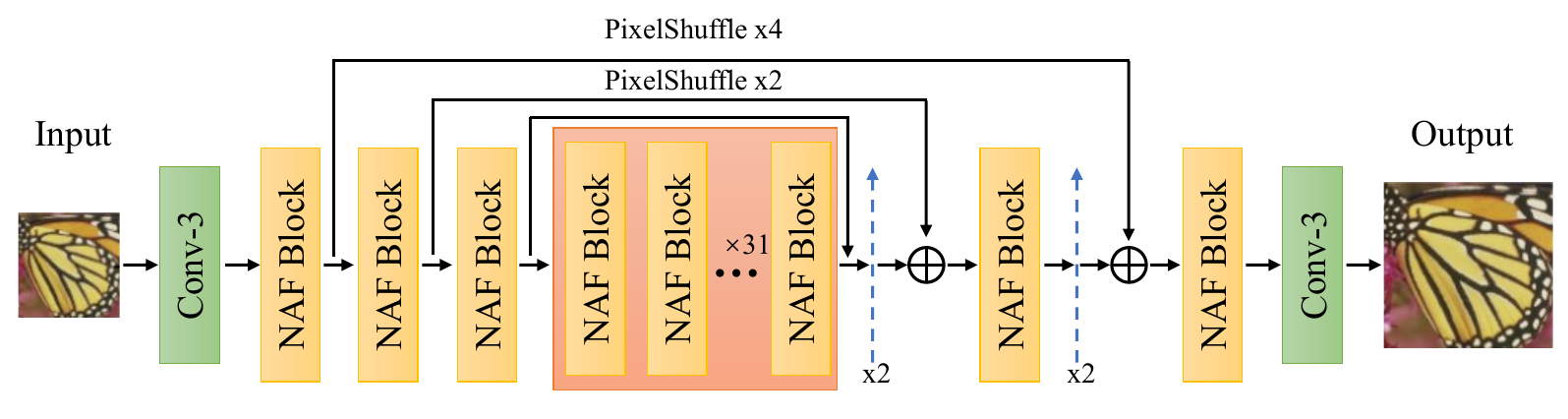}
		\caption{LVGroup\_HFUT: The structure of the proposed network. }
		\label{LVGroup_HFUT-1}
    \end{figure}

    \begin{figure}[t]
		\centering
		\includegraphics[width=\linewidth]{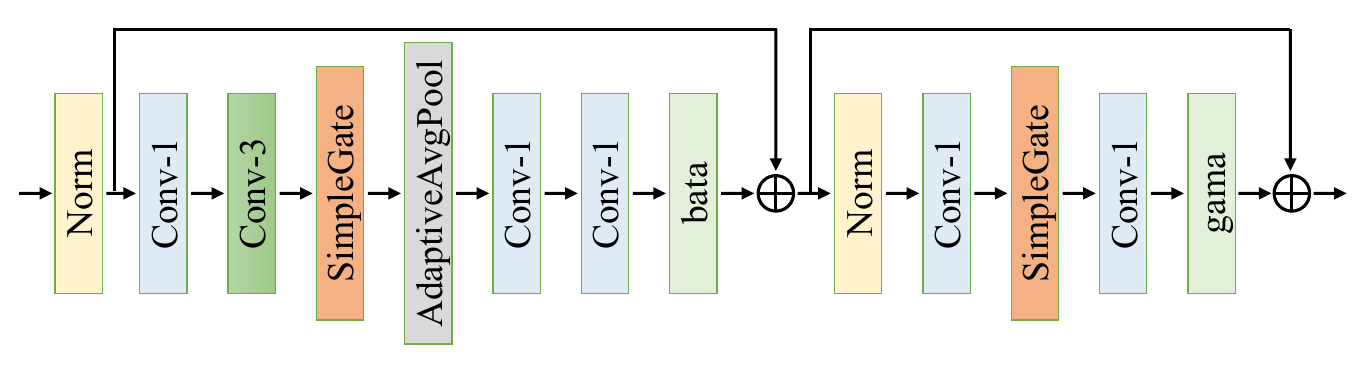}
		\caption{LVGroup\_HFUT: The structure of the proposed NAF block. }
		\label{LVGroup_HFUT-3}
    \end{figure}

     \subsection{DVision - Track 1}
     Despite numerous stereo image SR methods have been developed, these methods suffer huge computational cost. To meet the constraint of this challenge, this team analyzed the computational complexity of existing state-of-the-art stereo image SR models and introduced a improved version NAFSSR in Fig.~\ref{DVision}. 
     %Progressing with the state-of-the-art model, SwinFIRSSR is not viable due to its size surpassing the constraints. Consequently, they opted for the NAFSSR-T model, which, upon quantization, could adhere to the size limitations of the challenge. Choosing a smaller model entails a compromise in terms of perceptual quality. Therefore, their revised objective became implementing techniques that would yield visually pleasing output images while preserving the size of the trained model. This eliminated some options right off the bat, such as increasing the complexity of the model, both in terms of increasing the number of parameters and introducing complex architectures like transformers or diffusion-based models, which would lead to an expansion of parameters. Keeping all this in mind, we went ahead with the technique of Knowledge Distillation (KD).
     Specifically, the vanilla convolutional layers in NAFBlock were updated to omni-dimensional dynamic convolutions to achieve improved performance. 
     %The basic idea behind this approach is that they need rich and high-quality features that larger models generate. Still, they can’t use them directly but motivate our model (NAFSSRT) to generate outputs closer to what a larger model would do (SwinFIRSSR in our case). This technique induced a smaller model to generate outputs closer to what a larger and more complex model would generate. This led to a significant improvement in the perceptual quality of output SR images. The second novelty they brought to our framework is tweaking the NAFSSR algorithm’s underlying CNN architecture. The original implementation of the NAFSSR’s NAFBlock uses Conv2d layers. They replaced the last two of the five Conv2d layers of the NAFBlock’s architecture with ODConv2d, the Omni-dimensional Dynamic Convolution, which utilizes attention-based dynamic convolutions. We have observed significant improvements in the performance of CNN-based architectures when ODConv layers are used instead of conv2d layers. 

    \begin{figure*}
    \setcounter{figure}{25}
        \centering
        \includegraphics[width=1\linewidth]{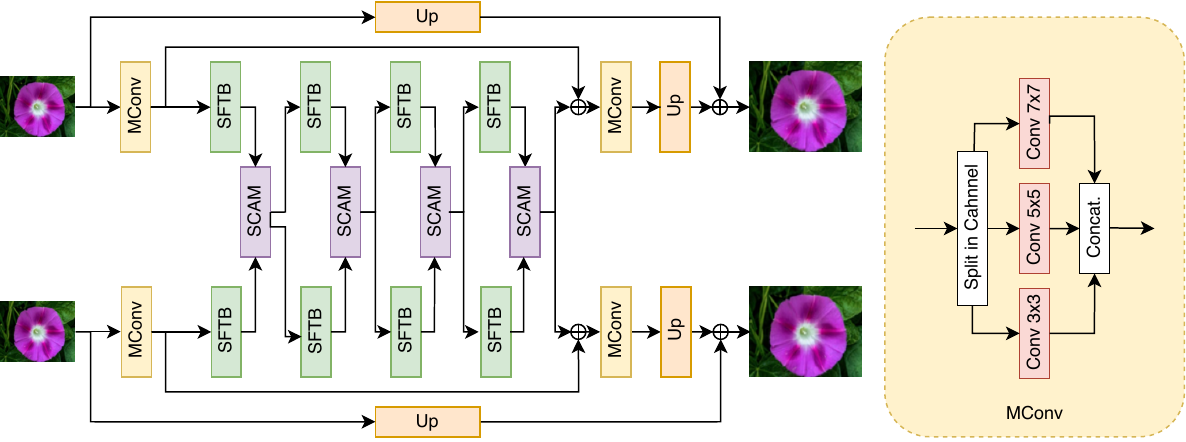}
        \caption{ECNU-IDEALab: The structure of the proposed network.}
        \label{fig:enter-label}
    \end{figure*}
    
    \begin{figure*}
        \centering
        \includegraphics[width=1\linewidth]{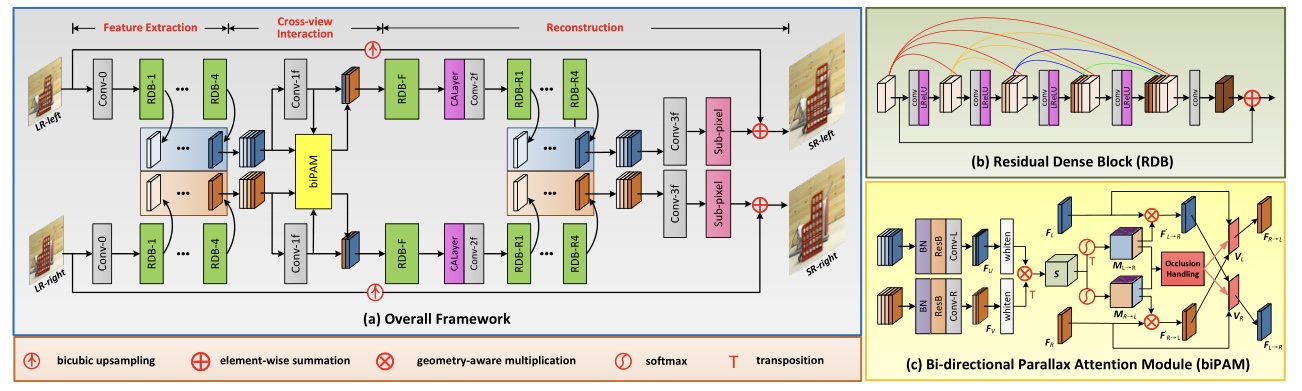}
        \caption{HiYun: The structure of the proposed iPASSR network.}
        \label{fig:iPASSRNetwork}
    \end{figure*}
    
    \subsection{LVGroup\_HFUT - Track 1, 2}

    This team employed NAFNet as their network architecture. The pipeline is shown in Fig.~\ref{LVGroup_HFUT-1} and the diagram is shown in Fig.~\ref{LVGroup_HFUT-3}. First, the input image is fed to a $3\times3$ convolution and cascaded residual feature blocks (RFB) for feature extraction. Then, bilinear interpolation upsampling as a mechanism to increase the spatial resolution of the feature maps. Concurrently, the upscaled features are passed through another convolutional layer for further refinement. Finally, these refined features are fed into a pixel shuffle layer to obtain the final results. 
    
    %During the training phase, the input image is fed into the NAFNet network, and it is constrained using three loss functions: lpips loss with weight 0.5, fft loss with weight 0.1, and L1 loss with weight 1.

    \textbf{Training Settings:} The proposed network was implemented on PyTorch 2.2.1 and an NVIDIA 4090 GPU. The proposed network was trained for 1500 epochs with a batch size of 32, using AdamW with $\beta_1=0.9$ and $\beta_2=0.999$ for optimization. The initial learning rate was set to 0.001. Random horizontal flip with probability 0.5 was used for data augmentation.

    \subsection{Liz620 - Track2}
    
    This team introduced a model named DconvNAFNet for stereo image SR. Specifically, the champion method (NAFSSR) \cite{Chu2022NAFSSR} in the NTIRE 2022 Challenge was used as the baseline. To construct a lightweight and efficient network with improved performance, a feature selective unit (DconvSCAM) based on depth-wise convolution was designed to aggregate different levels of features, as shown in Fig.~\ref{fig:DconvNAFSSR}. Multiple dataset augmentation strategies were employed, including RGB channel shuffling, horizontal/vertical shifting.  During the training process, the model was trained for 400k iterations on a NVIDIA 3090 GPU with a batch size of 16. The model was optimized using the Adam method with $ \beta_1$=0.9 and $\beta_2$=0.9. The initial learning rate was set to $ 2\times10^{-3} $, and the cosine annealing scheduler was used.

    \subsection{ECNU-IDEALab - Track2}
    
    This team proposed a multi-convolution parallel stereo image SR method, as shown in Fig.~\ref{fig:enter-label}. Based on SwinFIRSSR, the proposed MConv module is inserted to improve performance. Specifically, MConv divides the feature channels equally and then uses different convolution kernel sizes to extract multi-scale features. Finally, the features of different scales are concatenated along the channel dimension. %This can effectively improve model performance by increasing the model parameter count. In the final model, the MConv module uses convolution kernel sizes of 3, 5, and 7.

    \textbf{Training Settings:}
    During training, the images were cropped into $32\times32$ patches with a stride of 20. The proposed model was trained for 800k iterations with a batch size of 64. The learning rate was set to $2\times10^{-4}$ and halved at 600k, 650k, 700k, and 750k iterations. The CharbonnierLossColor loss and the Adam optimizer were employed for optimization.

    \subsection{HiYun - Track2}
    
    This team employed iPASSR as their network, which is illustrated in Fig.~\ref{fig:iPASSRNetwork}. The default settings of the original iPASSR were directly used for optimization.
    %After downloading the HR dataset from Codalab Track2 website, the high resolution images are downsized by 4 times respectively using the Matlab resize function in order to make training pairs with the corresponding high resolution images, then during the training session other parameters' settings are exactly the same as described in the "iPASSR" paper (previously mentioned and referenced). 

    \section{Acknowledgments}
    This work was partially supported by the National Natural Science Foundation of China (No. 62301601, U20A20185, 62372491, and 62301306), the Shenzhen Science and Technology Program (No. RCYX20200714114641140), and the Guangdong Basic and Applied Basic Research Foundation (2022B1515020103).
 This work was partially supported by the Humboldt Foundation. We thank the NTIRE 2024 sponsors: Meta Reality Labs, OPPO, KuaiShou, Huawei and University of W\"urzburg (Computer Vision Lab).
	
    \section{Organizers, Teams and Affiliations}
    \label{appendix}
	
    \subsection*{$\bullet$ NTIRE 2024 Organizers}
    \noindent \textbf{\textit{Challenge:}} NTIRE 2024 Challenge on Stereo Image Super-Resolution
	
    \noindent \textbf{\textit{Chairs:}} Longguang Wang$^1$, {Yulan Guo}$^{2,3}$, Yingqian Wang$^3$, Juncheng Li$^4$, Zhi Jin$^2$, Shuhang Gu$^5$, Radu Timofte$^6$
    
    \noindent \textbf{\textit{Technical Support:}} Hongda Liu$^2$, Yang Zhao$^{4}$
	
    \noindent \textbf{\textit{Affiliations:}} \\
	$^1$Aviation University of Air Force\\
	$^2$The Shenzhen Campus of Sun Yat-sen University, Sun Yat-sen University\\
	$^3$National University of Defense Technology\\
	$^4$Shanghai University\\
	$^5$University of Electronic Science and Technology of China\\
	$^6$Computer Vision Lab, University of W\"urzburg, Germany\\

    \subsection*{$\bullet$ NTIRE 2024 Teams}
    \subsection*{(1) Davinci - Track 1\textcolor[RGB]{255,215,0}{$^\bigstar$}, Track 2\textcolor[RGB]{255,215,0}{$^\bigstar$}}
    \noindent \textbf{\textit{Title:}} SwinFIRSSR: Stereo Image SR using SwinFIR
	
    \noindent \textbf{\textit{Members:}} Davinci (\textit{davinci7571@gmail.com}), Saining Zhang$^1$

    \noindent \textbf{\textit{Affiliations:}} \\
    $^1$Beijing Institute of Technology
    
    \subsection*{(2) HiSSR - Track 1\textcolor[RGB]{192,192,192}{$^\bigstar$}}

    \noindent \textbf{\textit{Title:}} RIISSR: Reference-based Iterative Interaction for Stereo Image SR
	
    \noindent \textbf{\textit{Members:}} Rongxin Liao$^1$ (\textit{2023110526@mail.hfut.edu.cn}), Ronghui Sheng$^2$, Feng Li$^1$, Huihui Bai$^2$, Wei Zhang$^3$, Runmin Cong$^3$

    \noindent \textbf{\textit{Affiliations:}} \\
    $^1$Hefei University of Technology\\
    $^2$Beijing Jiaotong University\\
    $^3$Shandong University

    \subsection*{(3) MiVideoSR - Track 1\textcolor[RGB]{184,115,51}{$^\bigstar$}, Track 2\textcolor[RGB]{192,192,192}{$^\bigstar$}}

    \noindent \textbf{\textit{Title:}} HCASSR
	
    \noindent \textbf{\textit{Members:}} Yuqiang Yang$^1$ (\textit{yangyuqiang@xiaomi.com}), Zhiming Zhang$^1$, Jingjing Yang$^1$, Long Bao$^1$, Heng Sun$^1$

    \noindent \textbf{\textit{Affiliations:}} \\
    $^1$Xiaomi Inc.
    
    \subsection*{(4) BUPTMM - Track 2\textcolor[RGB]{184,115,51}{$^\bigstar$}}

    \noindent\textbf{\textit{Title:}} Efficient CVHSSR
	
    \noindent \textbf{\textit{Members:}} Kanglun Zhao$^1$ (\textit{kanglunzhao@bupt.edu.cn}), Enyuan Zhang$^1$, Huiyuan Fu$^1$, Huadong Ma$^1$

    \noindent \textbf{\textit{Affiliations:}} \\
    $^1$Beijing University of Posts and Telecommunications
    
    \subsection*{(5) webbzhou - Track 1, Track 2}
    \noindent \textbf{\textit{Title:}} SOAN: Stereo Omnidirectional Aggregation Networks for Lightweight Stereo Image Super-Resolution

    \noindent \textbf{\textit{Members:}} Yuanbo Zhou$^1$ (\textit{webbozhou@gmail.com}), Wei Deng$^2$, Xintao Qiu, Tao Wang, Qinquan Gao, Tong Tong

    \noindent \textbf{\textit{Affiliations:}} \\
    $^1$Fuzhou University\\
    $^2$Imperial Vision Technology

    \subsection*{(6) Qi5 - Track 1}
    \noindent \textbf{\textit{Title:}} ESwinSSR

    \noindent \textbf{\textit{Members:}} Yinghao Zhu$^1$ (\textit{zhuyinghao0427@gmail.com}), Yongpeng Li$^1$
    
    \noindent \textbf{\textit{Affiliations:}} \\
    $^1$IQIYI

    \subsection*{(7) WITAILab - Track 1}
    \noindent \textbf{\textit{Title:}} CANSSR: Cross-view Aggregation Network for Stereo Image Super-resolution

    \noindent \textbf{\textit{Members:}} Zhitao Chen$^1$ (\textit{2290445045czt@gmail.com}), Xiujuan Lang$^1$, Kanghui Zhao$^1$, Bolin Zhu$^1$
    
    \noindent \textbf{\textit{Affiliations:}} \\
    $^1$Wuhan Institute of Technology

    \subsection*{(8) Giantpandacv - Track 1, Track 2}
    \noindent \textbf{\textit{Title:}} MIESSR: Efficient Multi-Level Information Extraction Network for Stereo Image Super-Resolution.

    \noindent \textbf{\textit{Members:}} Wenbin Zou$^1$ (\textit{alexzou14@foxmail.com}), Yunxiang Li$^2$, Qiaomu Wei$^3$, Tian Ye$^4$, Sixiang Chen$^4$
    
    \noindent \textbf{\textit{Affiliations:}} \\
    $^1$South China University of Technology\\
    $^2$Fuzhou University\\
    $^3$Chengdu University of Information Technology\\
    $^4$Hong Kong University of Science and Technology (Guangzhou)

    \subsection*{(9) JNU\_620 - Track 1, Track 2}
    \noindent \textbf{\textit{Title:}} Improved Loss for Stereo Image Super-Resolution Based on NAFSSR

    \noindent \textbf{\textit{Members:}} Weijun Yuan$^1$ (\textit{yweijun@stu2022.jnu.edu.cn}), Zhan Li$^1$, Wenqin Kuang$^1$, Ruijin Guan$^1$
    
    \noindent \textbf{\textit{Affiliations:}} \\
    $^1$Jinan University

    \subsection*{(10) GoodGame - Track 1, Track 2}
    \noindent \textbf{\textit{Title:}} CVHSSRPlus
	
    \noindent \textbf{\textit{Members:}} Jian Wang$^1$ (\textit{jwang4@snapchat.com}), Yuqi Miao$^2$, Baiang Li$^3$, Kejie Zhao$^4$

    \noindent \textbf{\textit{Affiliations:}} \\
    $^1$Snap Inc.\\
    $^2$Tongji University\\
    $^3$Hefei University of Technology\\
    $^4$Southern University of Science and Technology

    \subsection*{(11) Fly\_Flag - Track 1, Track 2}
    \noindent \textbf{\textit{Title:}} CPFSSR: Combined Permuted Self-Attention and Fast Fourier Convolution for Stereo Image Super-Resolution
	
    \noindent \textbf{\textit{Members:}} Wenwu Luo$^1$ (\textit{210220105@fzu.edu.cn}), Jing Wu$^1$

    \noindent \textbf{\textit{Affiliations:}} \\
    $^1$Fuzhou University

    \subsection*{(12) Mishka - Track 1, Track 2}
    \noindent \textbf{\textit{Title:}} VSSSR
	
    \noindent \textbf{\textit{Members:}} Yunkai Zhang$^1$ (\textit{1585832651@qq.com}), Songyan Zhang$^2$, Jingyi Zhang$^1$, Junyao Gao$^3$, Xueqiang You$^4$

    \noindent \textbf{\textit{Affiliations:}} \\
    $^1$Hefei University of Technology\\
    $^2$Nanyang Technological University\\
    $^3$Tongji University\\
    $^4$Zhuhai Zhongke HZ Technology Co.,Ltd

    \subsection*{(13) LightSSR - Track 1}
    \noindent \textbf{\textit{Title:}} LightSSR
	
    \noindent \textbf{\textit{Members:}} Yanhui Guo$^1$ (\textit{guoy143@mcmaster.ca}), Hao Xu$^1$

    \noindent \textbf{\textit{Affiliations:}} \\
    $^1$McMaster University

    \subsection*{(14) DVision - Track 1}
    \noindent \textbf{\textit{Title:}} NAFSSR-KD+ODConv
	
    \noindent \textbf{\textit{Members:}} Sahaj K. Mistry$^1$ (\textit{sahajmistry005@gmail.com}), Aryan Shukla$^1$, Sourav Saini$^1$, Aashray Gupta$^1$, Vinit Jakhetiya$^1$, Sunil Jaiswal$^2$.

    \noindent \textbf{\textit{Affiliations:}} \\
    $^1$Indian Institute of Technology Jammu\\
    $^2$K $\big|$ Lens GmbH

    \subsection*{(15) LVGroup\_HFUT - Track 1, Track 2}
    \noindent \textbf{\textit{Title:}} Light-NAFNet
    
    \noindent \textbf{\textit{Members:}} Zhao Zhang$^1$(\textit{cszzhang@gmail.com}), Bo Wang$^1$, Suiyi Zhao$^1$, Yan Luo$^1$, Yanyan Wei$^1$
    
    \noindent \textbf{\textit{Affiliations:}} \\
    $^1$Hefei University of Technology.
    
    \subsection*{(16) Liz620 - Track 2}
    \noindent \textbf{\textit{Title:}} DconvNAFNet
    
    \noindent \textbf{\textit{Members:}} Yihang Chen$^1$(\textit{Ehang@stu.jnu.edu.cn}), Ruting Deng$^1$, Yifan Deng$^1$
    
    \noindent \textbf{\textit{Affiliations:}} \\
    $^1$Jinan University.

    \subsection*{(17) ECNU-IDEALab - Track 2}
    \noindent \textbf{\textit{Title:}} MSFIRSSR
    
    \noindent \textbf{\textit{Members:}} Jingchao Wang$^1$(\textit{jcwang@stu.ecnu.edu.cn}), Zhijian Wu$^1$, Dingjiang Huang$^1$
    
    \noindent \textbf{\textit{Affiliations:}} \\
    $^1$East China Normal University.

    \subsection*{(18) HiYun - Track 2}
    \noindent \textbf{\textit{Title:}} iPASSR
    
    \noindent \textbf{\textit{Members:}} Yun Ye$^1$(\textit{ye.721@buckeyemail.osu.edu})
    
    \noindent \textbf{\textit{Affiliations:}} \\
    $^1$The Ohio State University.

	{\small
		\bibliographystyle{unsrt}
		\bibliography{super-resolution,other-CV-fields,neural-network,ntire_lhd,ntire24reports}
	}
	
\end{document}